\documentclass[letterpaper]{article} % DO NOT CHANGE THIS
\usepackage{aaai24}  % DO NOT CHANGE THIS
\usepackage{times}  % DO NOT CHANGE THIS
\usepackage{helvet}  % DO NOT CHANGE THIS
\usepackage{courier}  % DO NOT CHANGE THIS
\usepackage[hyphens]{url}  % DO NOT CHANGE THIS
\usepackage{graphicx} % DO NOT CHANGE THIS
\usepackage{natbib}  % DO NOT CHANGE THIS AND DO NOT ADD ANY OPTIONS TO IT
\usepackage{caption} % DO NOT CHANGE THIS AND DO NOT ADD ANY OPTIONS TO IT
\usepackage{algorithm}
\usepackage{algorithmic}

\usepackage{mathtools}
\usepackage{paralist}
\usepackage{subfigure}
\usepackage{url}            % simple URL typesetting
\usepackage{booktabs}       % professional-quality tables
\usepackage{amsfonts}       % blackboard math symbols
\usepackage{nicefrac}       % compact symbols for 1/2, etc.
\usepackage{microtype}      % microtypography
\usepackage{xcolor}         % colors
\usepackage{amsmath, amssymb}
\usepackage{bm}
\usepackage{newfloat}
\usepackage{listings}

\usepackage{multirow, makecell}

\DeclareCaptionStyle{ruled}{labelfont=normalfont,labelsep=colon,strut=off} % DO NOT CHANGE THIS
\floatstyle{ruled}
\newfloat{listing}{tb}{lst}{}
\floatname{listing}{Listing}
\title{Operator-learning-inspired Modeling of Neural Ordinary Differential Equations}
\author{
    Woojin Cho\textsuperscript{\rm 1}\equalcontrib,
    Seunghyeon Cho\textsuperscript{\rm 1}\equalcontrib,
    Hyundong Jin\textsuperscript{\rm 1},
    Jinsung Jeon\textsuperscript{\rm 1},\\
    Kookjin Lee\textsuperscript{\rm 2},
    Sanghyun Hong\textsuperscript{\rm 3},
    Dongeun Lee\textsuperscript{\rm 4},
    Jonghyun Choi\textsuperscript{\rm 1},
    Noseong Park\textsuperscript{\rm 1}
}
\affiliations{
    \textsuperscript{\rm 1}Yonsei University 
    \textsuperscript{\rm 2}Arizona State University 
    \textsuperscript{\rm 3}Oregon State University 
    \textsuperscript{\rm 4}Texas A\&M University-Commerce\\
    \{snowmoon, seunghyeoncho, tuzi04, jjsjjs0902\}@yonsei.ac.kr, \\ kjlee8344@gmail.com, sanghyun.hong@oregonstate.edu, dongeun.lee@tamuc.edu, \{jc, noseong\}@yonsei.ac.kr
}

\usepackage{bibentry}
\begin{document}

\maketitle

\begin{abstract}
Neural ordinary differential equations (NODEs), one of the most influential works of the differential equation-based deep learning, are to continuously generalize residual networks and opened a new field. They are currently utilized for various downstream tasks, e.g., image classification, time series classification, image generation, etc. Its key part is how to model the time-derivative of the hidden state, denoted $ \frac{d\mathbf{h}(t)}{dt}$. People have habitually used conventional neural network architectures, e.g., fully-connected layers followed by non-linear activations. In this paper, however, we present a neural operator-based method to define the time-derivative term. Neural operators were initially proposed to model the differential operator of partial differential equations (PDEs). Since the time-derivative of NODEs can be understood as a special type of the differential operator, our proposed method, called branched Fourier neural operator (BFNO), makes sense. In our experiments with general downstream tasks, our method significantly outperforms existing methods.
\end{abstract}

\section{Introduction}
% \label{submission}
\label{sec:intro}
Recently, in deep-learning community, there have been growing attention in developing differential-equation(DE)-based neural network architectures. One of the most representative DE-based models is neural ordinary differential equations (NODEs)~\cite{chen2018neural}, which serve as a general purpose neural network architecture, powered by deep implicit layers, modeling a continuous-depth representation and achieving state-of-the-art (SOTA) performance in many different down-stream tasks, e.g., time-series modeling~\cite{chen2018neural,song2020score,kidger2020neural,morrill2021neural,choi2021lt,jhin2022exit,kim2022sos,jeon2022gt}. Another fast growing area of deep learning is scientific machine learning (SML), where learning the deep-learning-based solution operator (also known as neural operators) of parametric partial differential equations has been of particular interest to researchers~\cite{li2020neural,cao2021choose,gupta2021multiwavelet,rahman2022u}. Fourier neural operators (FNOs)~\cite{li2020neural}, a representative class of neural operators, have demonstrated their effectiveness in building surrogate models for solution operators of parametric partial differential equations (PDEs). 

However, as opposed to NODEs, which have been explored in many different ML downstream tasks as generic neural network architectures, neural operators so far have been applied mostly to learning the solution operators of PDEs; the only exception we found is to use neural operators in computer vision task ~\cite{guibas2021adaptive}. Thus, in this study, we aim to bring the expressive power of neural operators into the modeling NODEs resulting in neural-operator-based NODE and apply this novel combination to many different downstream tasks. We are motivated to conduct this study based on a couple of observations: 
(1) as ODEs can be interpreted as a particular type of PDEs, which involve derivatives in only one variable, the ODE function can be modeled as neural operators, and 
(2) as neural operators are effective in learning ``parametric’’ PDE operators via a function-to-function mapping principle, the neural-operator-based NODE is expected to be equipped with stronger expressivity.    

\paragraph{Our approach:} NODEs model the dynamics of hidden state $\mathbf{h}(t)$ as a form of a system of ODEs from data such that %. The following ODE function, $f(\mathbf{h}(t), t; \boldsymbol{\theta}_{f})$ with learnable parameters $\boldsymbol{\theta}_{f}$, resides in their core parts:
% \begin{linenomath}
\begin{align}\label{eq:semi_NODE}
    \frac{d\mathbf{h}(t)}{dt} &=  f(\mathbf{h}(t), t; \boldsymbol{\theta}_{f}),
\end{align}
% \end{linenomath}
where $\boldsymbol{\theta}_{f}$ denotes a set of learnable parameters. Taking the forward pass is equivalent to solving the initial value problem (IVP) in Eq.~\eqref{eq:IVP} with the initial value $\mathbf{h}(t_{0})$:
% \begin{linenomath}
\begin{align}\label{eq:IVP}
    \mathbf{h}(t_{1}) = \mathbf{h}(t_{0}) + \int_{t_0}^{t_1} f(\mathbf{h}(t), t; \boldsymbol{\theta}_{f}) dt,
\end{align}
% \end{linenomath}
and the last hidden state $\mathbf{h}(t_{1})$ is typically used for downstream tasks (e.g., classification). 

The ODE function (the right hand side, or RHS) in Eq.~\eqref{eq:semi_NODE} can be considered as applying the differential operator to $\mathbf{h}(t)$ and it is natural to redesign RHS, $f$, as neural operators. From our empirical studies, we observed that na\"ively adapting recent neural operator methods, such as Fourier neural operators (FNOs), to NODEs, however, results in degradation of performance. We speculate that this degradation is caused by a habitual choice of a rather simple architectural design, which uses conventional neural networks, (e.g., convolutional neural networks (CNNs) and fully-connected networks (FCNs)) to learn the differential operator. We, instead, propose a novel neural operator architecture called branched Fourier neural operators (BFNOs), which leads to learning of much more expressive neural operators that can be applied to complex ML downstream tasks.

For empirical evaluations, we test our method on various ML downstream tasks including image classification, time series classification, and image generation. For each task, we compare our method with SOTA NODE and non-NODE-based baselines. In particular, our main competitors are enhanced NODE models, such as heavy ball NODEs (HBNODEs). Our BFNO-based NODEs (BFNO-NODEs) outperforms all %those advanced NODE models and some other non-NODE-based 
considered baselines. We summarize our contributions as follows:

\begin{compactenum}
\item \textbf{Branched Fourier neural operator:} We design the ODE function of NODEs using our proposed BFNO method. %Our BFNO is specialized for being used as ODE functions. 
In our design, we do not inherit previous architectures of FNO variants, i.e., applying a low-pass filter to the Fourier transform output, followed by a convolution operation, as it is rather restrictive to be used for general machine learning tasks. Instead, our BFNO uses dynamic global convolutional operations with multiple kernels. To our knowledge, we are the first using neural operators to design the ODE function of NODEs.
% We build our model using neural operator and Neural ODEs, and we adopt modified Fourier Neural Operator (FNO) as an effective ODE function. Unlike PDE tasks where low frequency data is important, high frequency data is also important in general machine learning tasks. For example, images have high-resolution content with edges and other structures. Therefore, we need not only low frequency data but also high frequency data. However, FNO consists of too many weights to consider all frequencies. Our \solution{} uses the Branched Fourier Neural Operator (BFNO) that revises the Fourier layer of FNO to reduce the number of weights and allows us to consider all of the frequencies.
% \item \textbf{Continuous Width Neural ODEs :} Neural ODEs are a continuous-depth neural network. We design  \solution{} using neural operator. Due to the function to function characteristics of neural operator, Neural ODEs are added a continuous width property. In other words, our \solution{} are a continuous depth and width neural network. Our \solution{} have the advantage of reducing the computation. Even if the resolutions of the training data and the inference data mismatch, our model can still function.
% \,
% \,
\item \textbf{High performance in general ML tasks:} Through extensive experiments, we demonstrate that BFNO-NODEs significantly outperform considered baselines in various ML downstream tasks. Our experiments set include image classification, time series classification, and image generation. Our method improves on the performance of previous SOTA NODE and non-NODE methods by 20\% in image classification, 1.7\% in time series classification in terms of test accuracy, and 10\% in image generation (in terms of the standard metrics for those tasks).
% {\color{red}{Furthermore, our model confirms the result of less overfitting than the existing NODE. Even if the input resolution increases, our model does not have computation occurring externally, but the existing NODE requires a lot of computation.}} 7% 3% 9.2%
\end{compactenum}

\section{Related work}
\label{sec:related_work}

\paragraph{Continuous-depth neural networks:} Continuous-depth neural networks allow greater modeling flexibility and have been investigated as a potential replacement for traditional deep feed-forward neural networks~\cite{weinan2017proposal,haber2017stable,lu2018beyond}. NODEs~\cite{chen2018neural} can model continuous-depth neural networks. By the continuous-depth property, NODEs describe the change of the hidden state $\mathbf{h}(t)$ over time. In particular, NODEs use the adjoint sensitivity method to address the main technical challenge of training continuous-depth networks. Many continuous-depth models that use the same computational formalism, such as neural controlled differential equations~\cite{kidger2020neural}, neural stochastic differential equations~\cite{liu2019neural}, and so on, have been proposed. In addition, NODE-based models with enhanced ODE function architectures have achieved significant improvements in various tasks in comparison with the original NODE design. Augmented NODEs (ANODEs~\cite{dupont2019augmented}) suggested solving the homeomorphic limitation of ODEs with adding extra dimensions. Second-order NODEs (SONODEs~\cite{norcliffe2020second}) showed that they can expand the adjoint sensitivity approach to the second-order ODEs efficiently. In order to further enhance the training and inference of NODEs, heavy ball NODEs (HBNODEs~\cite{xia2021heavy}) model the dynamics with the conventional momentum method. Nesterov NODEs (NesterovNODEs~\cite{nguyenimproving}) is proposed to use the Nesterov momentum expression with a fast convergence rate among optimizer methods. Adaptive moment estimation NODEs (AdamNODEs~\cite{cho2022adamnodes}) use an enhanced momentum-based method to define the ODE function. In this paper, we propose a novel continuous-depth NODE architecture with the infinite dimensional property %based on our proposed BFNO. 
inherited from characteristics of neural operators. 
Our enhanced NODE architecture, called BFNO-NODE, outperforms existing NODE enhancements.

\paragraph{Operator learning:}
Expressing and analyzing changes in phenomena over time and space as PDEs are an important issue in natural science and engineering. However, solving PDE problems is a mathematically difficult task. In the field of numerical analysis, these problems are approached using various numerical methods~\cite{smith1985numerical,reddy2019introduction}. These methods perform well in many areas where PDE problems need to be analyzed, but there are fundamental challenges: i) it takes a lot of computational cost to solve the problem, ii) the higher its required accuracy, the longer its computation time, and iii) finally, it is difficult to analyze for various spatial resolutions with a single model.

To overcome these limitations, neural operators~\cite{lu2019deeponet,kovachki2021neural} have recently been proposed in the field of deep learning. Neural operators aim to learn the (inverse) differential operator of PDEs. Furthermore, they also have the infinite dimensional property.

Recently, models with repetitive operator-based layers have been proposed in the field of neural operators~\cite{li2020neural,li2020multipole,wen2022u}. Typically, FNOs first define a kernel integral operator layer in the Fourier domain and repeatedly stack multiple such layers. In the field of computer vision, there was a study that improved the vision transformer~\cite{dosovitskiy2020image} using FNOs. Adaptive Fourier neural operator (AFNO) was proposed in~\cite{guibas2021adaptive} by replacing the self-attention layers of the vision transformer with  adaptive FNO layers. However, these operator architectures did not perform well when they were used in the ODE function in our preliminary experiments. Therefore, we propose our BFNO concept.

\begin{figure*}[t]
\centering
% \vspace{-1.5em}
\includegraphics[width=1.7\columnwidth]{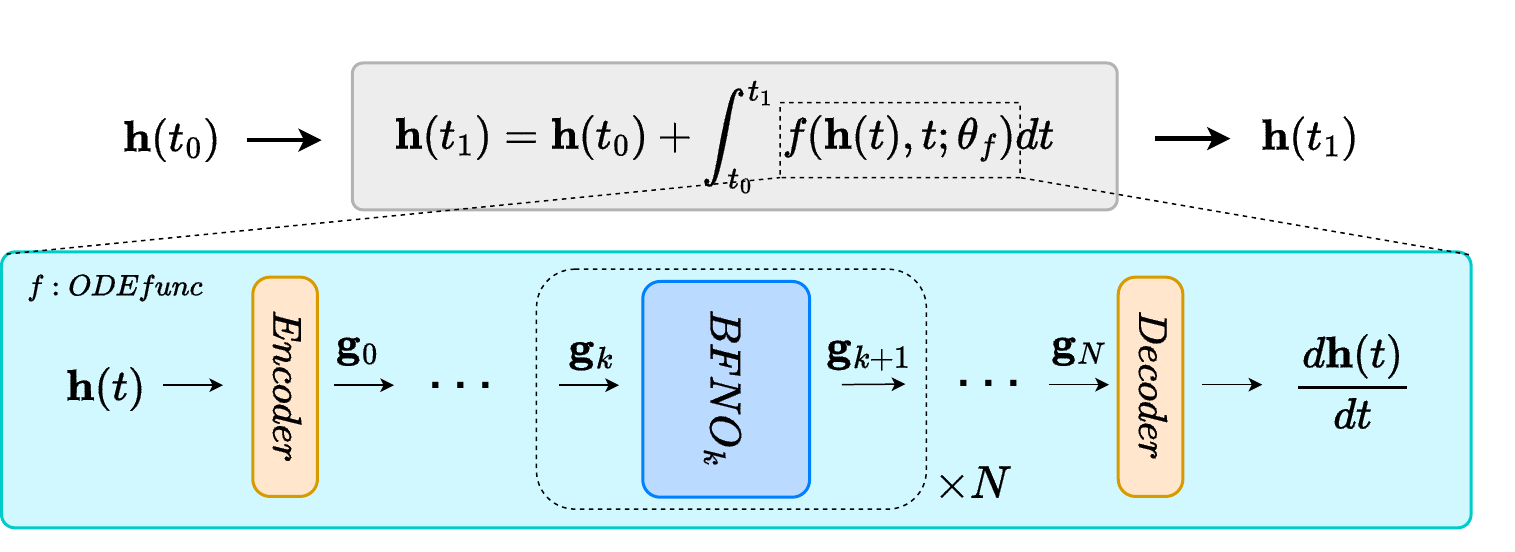}
\caption{The overall architecture of BFNO-NODE. Our goal is to define the ODE function $f$ based on our proposed neural operator.}

% \vspace{-1em}
\label{fig:main_figure}
\end{figure*}

\section{Preliminaries}

\paragraph{Neural ordinary differential equations (NODEs):}
In order to determine $\mathbf{h}(t_{1})$ from $\mathbf{h}(t_{0})$, NODEs solve the integral problem in Eq.~\eqref{eq:IVP}, where the ODE function $f(\mathbf{h}(t), t; \boldsymbol{\theta}_{f})$ is a neural network used to approximate $\frac{d\mathbf{h}(t)}{dt}$ (cf. Eq.~\eqref{eq:semi_NODE}). Once $f$ is trained, NODEs rely on numerical ODE solvers to solve the integral problem, such as the explicit Euler method, the Dormand-Prince (DOPRI~\cite{DORMAND198019}) method. ODE solvers discretize the time variable $t$ and transform an integral into numerous additions. One distinguished characteristic of NODEs is that the gradient of loss w.r.t. $\boldsymbol{\theta}_{f}$, denoted $ \nabla_{\boldsymbol{\theta}_{f}}C = \frac{\partial C}{\partial \boldsymbol{\theta}_{f}}$, where  $C$ is a task-dependent cost function, can be calculated by a reverse-mode integration, which has $\mathcal{O}(1)$ space complexity. This gradient calculation method is known as the adjoint sensitivity method.

\paragraph{Operator learning and discretization of function:}
Let $S$ be a bounded, open set which is a subspace of $\mathbb{R}^{d}$, and let $P=P(S; \mathbb{R}^{d_p})$ and $Q=Q(S; \mathbb{R}^{d_q})$ be Banach spaces which define a function mapping from input $x$ $\in$ $S$ to output whose dimensions are $d_p$ and $d_q$ respectively. Then, a non-linear operator $\mathcal{L}$ : $P$ $\xrightarrow{}$ $Q$, where $P$ and $Q$ are the continuous function spaces, can be defined. By discretizing with finite numbers of observations, the function spaces can be represented in grid forms. At the end, neural operators learn a parameterized mapping $\mathcal{L_\theta}$ : $P$ $\xrightarrow{}$ $Q$ such that:
%\begin{linenomath}
\begin{align}
    {\mathcal{L_\theta} (P)(x)} = Q(x),  \qquad  {\forall{x}} \in {S}.
    \label{eq:operator_eq}
\end{align}
%\end{linenomath}

For the given two functions, $P(x)$ and $Q(x)$ with ${x} \in {S}$, the operator $\mathcal{L}_{\theta}$ can map these function spaces. As such, NODE's time-derivative operator $\mathcal{D}{_f} = \frac{d{(\cdot)}} {dt}$ can be understood as a function mapping from $\mathbf{h}(t)$ to $\frac{d{\mathbf{h}}(t)} {dt}$, where $t$ denotes the temporal coordinate. This differential operator can be regarded as a special case of the general operator representation. Therefore, we can design the ODE function $f$ of NODEs as an operator-based network.

\paragraph{Kernel integral operators:} As a way of expressing an operator, one can define the kernel integral operator $\mathcal{K}$ as following:
% \begin{linenomath}
\begin{align}
    % {\mathcal{K}(a;\psi){g_k}}(x) := \int_D {\kappa(x, /y, a(x), a(y);\psi){{g_k}(y)}}dy
    {\mathcal{K}{(q)}}(x) := \int_S {\kappa(x, y) {q(y)}} dy,   \qquad   {\forall{x}} \in {S}, 
    \label{eq:kernel_integral_eq}
\end{align}
% \end{linenomath}
where $\kappa$ is a continuous kernel function. If we use the Green's kernel $\kappa(x, y) = \kappa(x-y)$, the RHS of Eq.~\eqref{eq:kernel_integral_eq} becomes the following global convolution operator:
% \begin{linenomath}
\begin{align}
    % {\mathcal{K}(a;\psi){g_k}}(x) := \int_D {\kappa(x, y, a(x), a(y);\psi){{g_k}(y)}}dy
    %{\mathcal{K}{(q)}}(x) := \int_S {\kappa(x-y) {q(y)}} dy, \qquad   {\forall{x}} \in {S}.
    \int_S {\kappa(x-y) {q(y)}} dy, \qquad   {\forall{x}} \in {S}.
    \label{eq:global_conv_eq}
\end{align}
% \end{linenomath}

% \begin{align}
%     % {\mathcal{K}(a;\psi){g_k}}(x) := \int_D {\kappa(x, y, a(x), a(y);\psi){{g_k}(y)}}dy
%     {\mathcal{K}{(\alpha)}}(x) := \int_D {\kappa(x, y;\psi) {{\alpha}(y)}} dy,
%     \label{eq:kernel_integral_eq}
% \end{align}

% where ${\kappa}_\psi : \mathbb{R}^{2(d+d_a)} \xrightarrow{} \mathbb{R}^{d_v \cdot d_v}$ is a neural network parameterized by $\psi \in {\boldsymbol{\theta}}_{\kappa}$, and input tensor $\alpha$ is infinite-dimensional function defined on domain $D \subset \mathbb{R}^2$. If ${\kappa}_{\psi}(x, y)$ becomes ${\kappa}_{\psi}(x-y)$, Equation~\ref{eq:kernel_integral_eq} can be defined as convolution operator~\cite{li2020fourier}.   

\paragraph{Fourier neural operators (FNOs):}
For an efficient parameterization of the kernel $\kappa$, FNOs rely on the Fourier transform as a projection of a function onto the Fourier domain~\cite{li2020fourier}. Given an input function $\mathbf{g}(x)$, for instance, FNOs use the following kernel integral operator $\mathcal{K}$ parameterized by $\psi$~\cite{li2020fourier}:
% \begin{linenomath}
\begin{align}
    % {(\mathcal{K}(\psi){g_k})}(x) = \mathcal{F}^{-1}(R \cdot \mathcal{F}(g_k))(x)
    {(\mathcal{K}(\psi)\mathbf{g})}(x) = \mathcal{F}^{-1}(\mathbf{R}_\psi \odot \mathcal{F}(\mathbf{g}))(x), \qquad   {\forall{x}} \in {S},
    \label{eq:kernel_fourier}
\end{align}
% \end{linenomath}

where $\mathcal{F}$ denotes the fast Fourier transform and $\mathcal{F}^{-1}$ its inverse. $\odot$ denotes the elementwise multiplication, and $\mathbf{R}_\psi$ denotes a tensor representing a global convolutional kernel. Note that the global convolutional kernel $\mathbf{R}_\psi$ is the only trainable parameter, i.e., $\psi \equiv \mathbf{R}_\psi$.

\begin{figure*}[t]
\centering
\subfigure[BFNO layer]{\includegraphics[width=0.56\columnwidth]{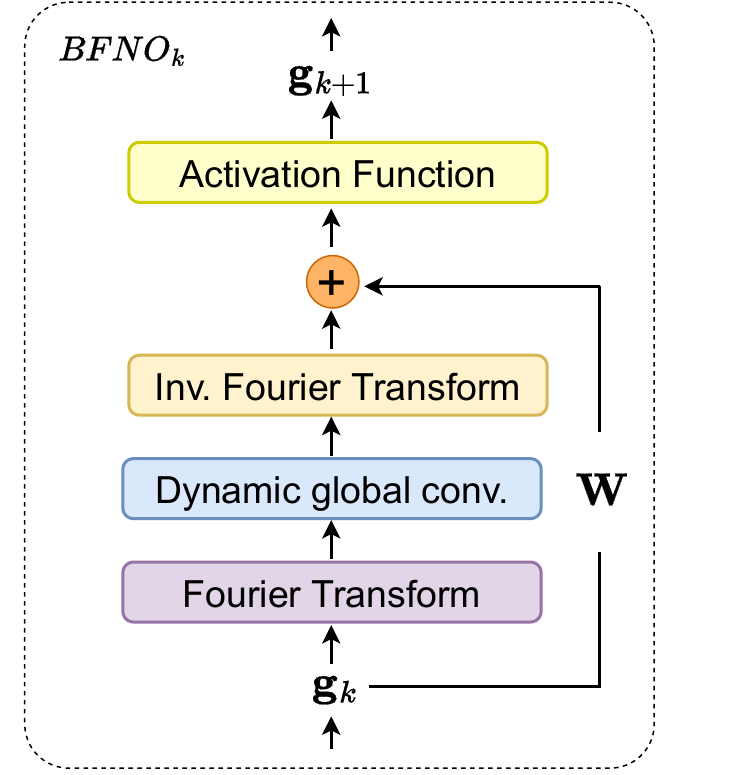}}
\subfigure[Dynamic global convolution]{\includegraphics[width=1.40\columnwidth]{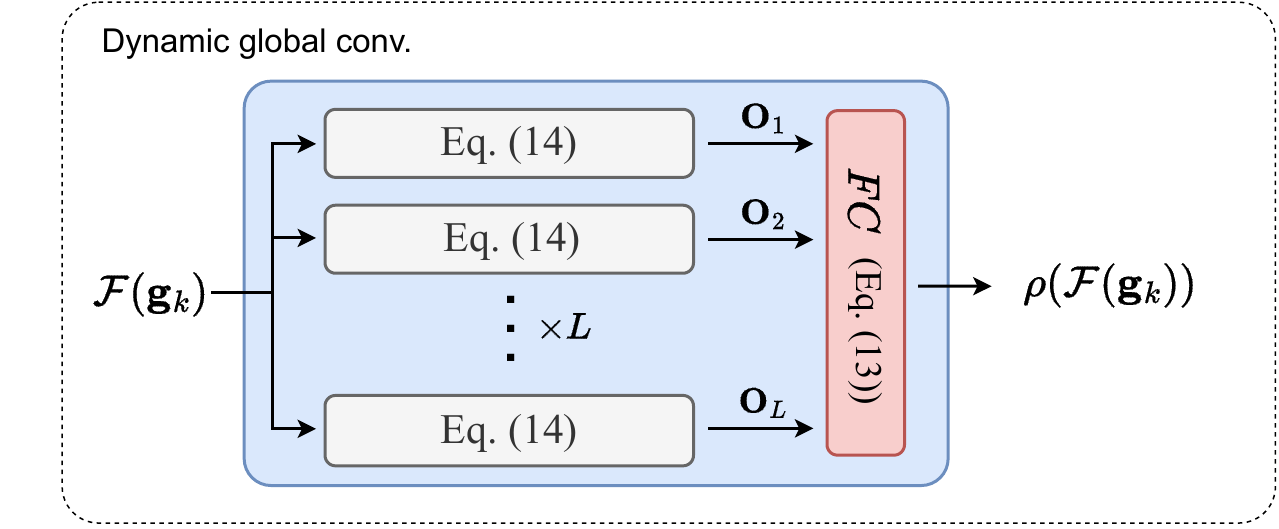}}
% \vspace{-1em}
\caption{The detailed proposed architecture. We perform the Fourier transform, the dynamic global convolutional operation, and the inverse Fourier transform to process $\mathbf{g}_{k}$ in conjunction with one additional transformation by the transformation matrix $\mathbf{W}$ 
% (cf. Eq.~\eqref{eq:real_imp}).
.}
% \vspace{-1em}
\label{fig:detail_architecture}
\end{figure*}

\section{Proposed method}
\label{sec:CWNODE}
In this section, we describe our motivations, followed by the overall workflow and detailed model architecture.

\subsection{Motivations}
The ODE function $f$ in Eq.~\eqref{eq:semi_NODE} is to learn the time-derivative operator of the hidden state $\mathbf{h}(t)$, in which case we can consider that $f$ is a special operator suitable for defining NODEs. Therefore, our proposed NODE concept can be rewritten as follows:
% In Equation~\ref{eq:operator_eq}, The differential operator $\mathcal{D}$ that Neural Operator wants to learn is the change in time and space of PDE. If $\mathcal{D}$ is a differential operator for time, it becomes the same as \Eqref{eq:semi_NODE}. Eventually, \Eqref{eq:full_NODE} can be transformed as follows by applying time derivative operator $\mathcal{D}_f$, a special case of differential operator $\mathcal{D}$.
% \begin{linenomath}
\begin{align}
    \mathbf{h}(t_1) &= \mathbf{h}(t_0) + \int_{t_0}^{t_1} {\mathcal{D}_f(\mathbf{h}(t), t; \boldsymbol{\theta}_{f})} \mathrm dt,
    \label{eq:operator_NODE}
\end{align}
% \end{linenomath}

where $\mathcal{D}_f$ is the time-derivative operator. However, we found that FNOs (Eq.~\eqref{eq:kernel_fourier}) are not suitable for use in general machine learning tasks (see Section~\ref{ablation} for detailed discussion).
Therefore, we propose our own extended FNO method, called branched Fourier neural operator (BFNO).

Many NODE-based papers have habitually used conventional architectures, e.g., fully-connected layers, convolutional layers, and rectified linear units (ReLUs), to define the ODE function~\cite{dupont2019augmented,norcliffe2020second,xia2021heavy,nguyenimproving,cho2022adamnodes}. In comparison with them, our method provides a more rigorous way for defining the ODE function (or the time-derivative operator). Our main goal in this paper is to enhance the model accuracy by learning the ODE function more accurately.

\subsection{Overall workflow}
The overall model architecture of our proposed BFNO-NODE is shown in Figure~\ref{fig:main_figure}. There are three major points in our design in contrast to the original NODE design: i) the neural operator-based ODE function, ii) the BFNO layer, and iii) the dynamic global convolution. We first sketch the overall mechanism before describing details as follows:
\begin{compactenum}
\item Given $\mathbf{h}(t)$ and $t$, we are interested in calculating $\frac{d\mathbf{h}(t)}{dt}$. Its first step is encoding $\mathbf{h}(t)$ and $t$ into a representation $\mathbf{g}_0$.
\item A series of $N$ BFNO layers follows to calculate $\mathbf{g}_N$, which will be decoded into $\frac{d\mathbf{h}(t)}{dt}$. Therefore, the key in our design is how the BFNO layer works.

\begin{compactenum}
    \item In the $k$-th BFNO layer shown in Figure~\ref{fig:detail_architecture} (a), the first step is to apply the Fourier transform to its input $\mathbf{g}_k$, followed by a dynamic global convolution and the inverse Fourier transform.
    \item According to the convolution theorem~\cite{bracewell1986fourier}, the global convolution becomes an elementwise multiplication with a global filter in the Fourier domain (cf. Eq.~\eqref{eq:dg2}). Our dynamic global convolution i) applies $L$ different filters, and ii) aggregates those $L$ different outcomes with a fully-connected layer (cf. Eq.~\eqref{eq:dg1}) as shown in Figure~\ref{fig:detail_architecture} (b).
    \item There is another processing path for applying the linear transformation $\mathbf{W}$ to $\mathbf{g}_k$.
    \item The previous two results are added, and a non-linear activation function finally outputs $\mathbf{g}_{k+1}$.
\end{compactenum}
\end{compactenum}

\subsection{BFNO-NODE}% Model Architecture}
Now, we describe our proposed method in detail.

\paragraph{Neural operator-based ODE functions:} We propose the following ODE function with our newly designed BFNO layer:
% \begin{linenomath}
\begin{align}\begin{split}
    \mathbf{g}_0 &= Encoder(\mathbf{h}(t), t),\\
    \mathbf{g}_{k+1} &= BFNO_k(\mathbf{g}_k), 0 \leq k \leq N-1,\\
    \frac{d\mathbf{h}(t)}{dt} &= Decoder(\mathbf{g}_N),
\end{split}\end{align}
% \end{linenomath}
where $BFNO_k$ is the $k$-th BFNO layer. Both the time-augmented encoder ($Encoder$) and the decoder ($Decoder$) consist of fully-connected layers. The size of encoded vector $\mathbf{g}_k$, $0 \leq k \leq N$, is a hyperparameter. We use $\dim(\mathbf{g})$ to denote the size of $\mathbf{g}_k$. The output size of the decoder is the size of $\mathbf{h}(t)$. The size of $\mathbf{h}(t)$, denoted $\dim(\mathbf{h})$, is also a hyperparameter.

\paragraph{Branched Fourier neural operator (BFNO) layers:} Figure~\ref{fig:detail_architecture} (a) shows the structure of the BFNO layer. The update process between the input value $\mathbf{g}_{k}$ and the output value $\mathbf{g}_{k+1}$ of the $k$-th BFNO layer is expressed as follows:
% \begin{linenomath}
\begin{align}
{\mathbf{g}_{k+1}}(x) &= BFNO_k\big(\mathbf{g}_{k}(x)\big)\\
     &\Rightarrow \sigma\big((\mathcal{K}(\mathbf{h},t;\psi)\mathbf{g}_{k})(x) + {\mathbf{W}}{\mathbf{g}_{k}}(x)\big)\\
    &\Rightarrow \sigma\big(\mathcal{F}^{-1}(\rho(\mathcal{F}(\mathbf{g}_{k})))(x) + {\mathbf{W}}\mathbf{g}_{k}(x)\big),\label{eq:update_eq}
\end{align}
% \end{linenomath}
where the first term is a kernel integral operator parameterized by $\psi$, and the second term is a linear transformation parameterized by $\mathbf{W}$. $\sigma$ is an activation function. $\rho$ is the dynamic global convolutional operation for the input $\mathbf{g}_k$, which will be described shortly.

Under the regime of our proposed BFNO, $\mathbf{g}$ is theoretically a spatial function. However, infinite-dimensional spatial functions cannot be processed by modern computers and therefore, all neural operator methods implement their discretized versions, which is also the case for our method. As a result, our actual implementations are as follows:
% \begin{linenomath}
\begin{align}
    {\mathbf{g}_{k+1}} = \sigma\big(\mathcal{F}^{-1}(\rho(\mathcal{F}(\mathbf{g}_{k}))) + \mathbf{W} {\mathbf{g}_{k}}\big),
\label{eq:real_imp}
\end{align}
% \end{linenomath}
where $\mathbf{g}_k$ (resp. $\mathbf{g}_{k+1})$ is an input (resp. output) vector. $\mathbf{W} \in \mathbb{R}^{\dim(\mathbf{g}) \times \dim(\mathbf{g})}$ is a linear transformation matrix, whose sizes are square matrix with the same number of rows and columns.

\begin{table*}[t]
\centering
\setlength{\tabcolsep}{2pt}
\footnotesize
\setlength{\tabcolsep}{2pt}
\renewcommand{\arraystretch}{1.2}
% \vspace{.5em}
\begin{tabular}{lcccccccccc}
\specialrule{1pt}{1pt}{1pt}
\multirow{2}{*}{Method} & \multicolumn{4}{c}{Number of parameters}& & \multicolumn{4}{c}{Test Accuracy} \\ \cline{2-5} \cline{7-10}
 &  MNIST & CIFAR-10 & CIFAR-100 & STL-10 & & MNIST & CIFAR-10 & CIFAR-100 & STL-10  \\ \hline
NODE & 85,316 & 173,611 & 646,021  & 521,512 & & 0.9531 $\pm$ 0.0042  & 0.5466 $\pm$ 0.0051  & 0.2178 $\pm$ 0.0024  & 0.3469 $\pm$ 0.0087  \\
ANODE & 85,462 & 172,452 & 645,121 & 520,180 & & 0.9816 $\pm$ 0.0024 & 0.6025 $\pm$ 0.0032  & 0.2566 $\pm$ 0.0052 & 0.3711 $\pm$ 0.0089  \\ 
SONODE & 86,179 & 171,635 & 645,081 & 521,532 & & 0.9824 $\pm$ 0.0013  & 0.6132 $\pm$ 0.0073  & 0.2591 $\pm$ 0.0058 & 0.3646 $\pm$ 0.0091  \\
HBNODE & 85,931 & 172,916 & 646,338 & 521,248 & & 0.9814 $\pm$ 0.0011  & 0.5989 $\pm$ 0.0035  & 0.2549 $\pm$ 0.0057  & 0.3426 $\pm$ 0.0036  \\ 
GHBNODE & 85,931 & 172,916 & 646,338 & 521,248 & & 0.9817 $\pm$ 0.0005  & 0.6085 $\pm$ 0.0050  & 0.2608 $\pm$ 0.0076  & 0.3648 $\pm$ 0.0078  \\   
NesterovNODE & 85,930 & 172,915 & 647,869 & 523,351 & & 0.9824 $\pm$ 0.0015  & 0.5996 $\pm$ 0.0033  & 0.2508 $\pm$ 0.0089  & 0.3595 $\pm$ 0.0051  \\
GNesterovNODE & 85,930 & 172,915 & 647,869 & 523,351 & & 0.9807 $\pm$ 0.0013 & 0.6172 $\pm$ 0.0064  & 0.2568 $\pm$ 0.0060  & 0.3690 $\pm$ 0.0091  \\
AdamNODE & 85,832 & 171,904 & 644,974 & 521,192 & & \textbf{0.9834 $\pm$ 0.0004}  & 0.6264 $\pm$ 0.0015  & 0.2405 $\pm$ 0.0049  & 0.3606 $\pm$ 0.0019   \\ \hline
BFNO-NODE & 85,591 & 171,333 & 644,547 & 514,291 & & 0.9752 $\pm$ 0.0021  & \textbf{0.6289 $\pm$ 0.0054} & \textbf{0.2890 $\pm$ 0.0094}  & \textbf{0.4455 $\pm$ 0.0029}   \\ 
% & FFJORD ORIGIN & CIFAR-10  \\ \hline
% MNIST       & 0.99 & 3.40             \\ 
% CIFAR-10           & 0.97 & 3.38    \\   \hline
% FFJORD CWNODE (OURS)   &\textbf{0.88}         &      \textbf{3.33}               \\
\specialrule{1pt}{1pt}{1pt}
\end{tabular}
\caption{The size of models and the test accuracy for MNIST, CIFAR-10, CIFAR-100, and STL-10 image classification. Our BFNO-based NODEs outperform other methods. Each baseline method has its own design guidance for the ODE function $f$ as in our method.}
\label{tab:Imageclassification_param_acc}
\end{table*}

\begin{figure*}[t]
    \centering
    % \vspace{-1.5em}
    \includegraphics[width=1\textwidth]{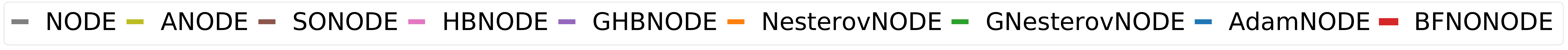} %\vspace{-1em}
    \subfigure[][MNIST]{\includegraphics[width=0.245\textwidth]{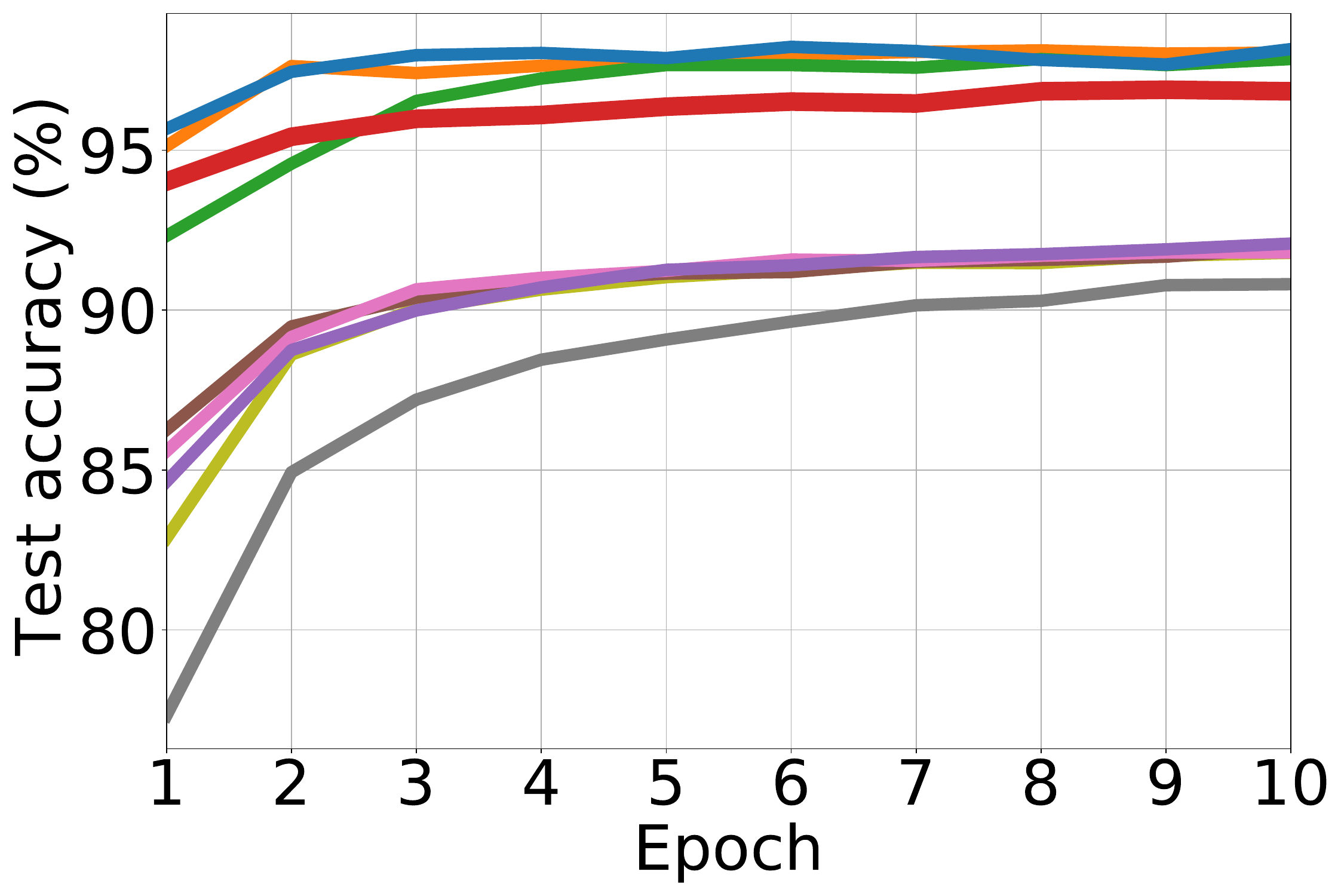}}
    \subfigure[][CIFAR-10]{\includegraphics[width=0.245\textwidth]{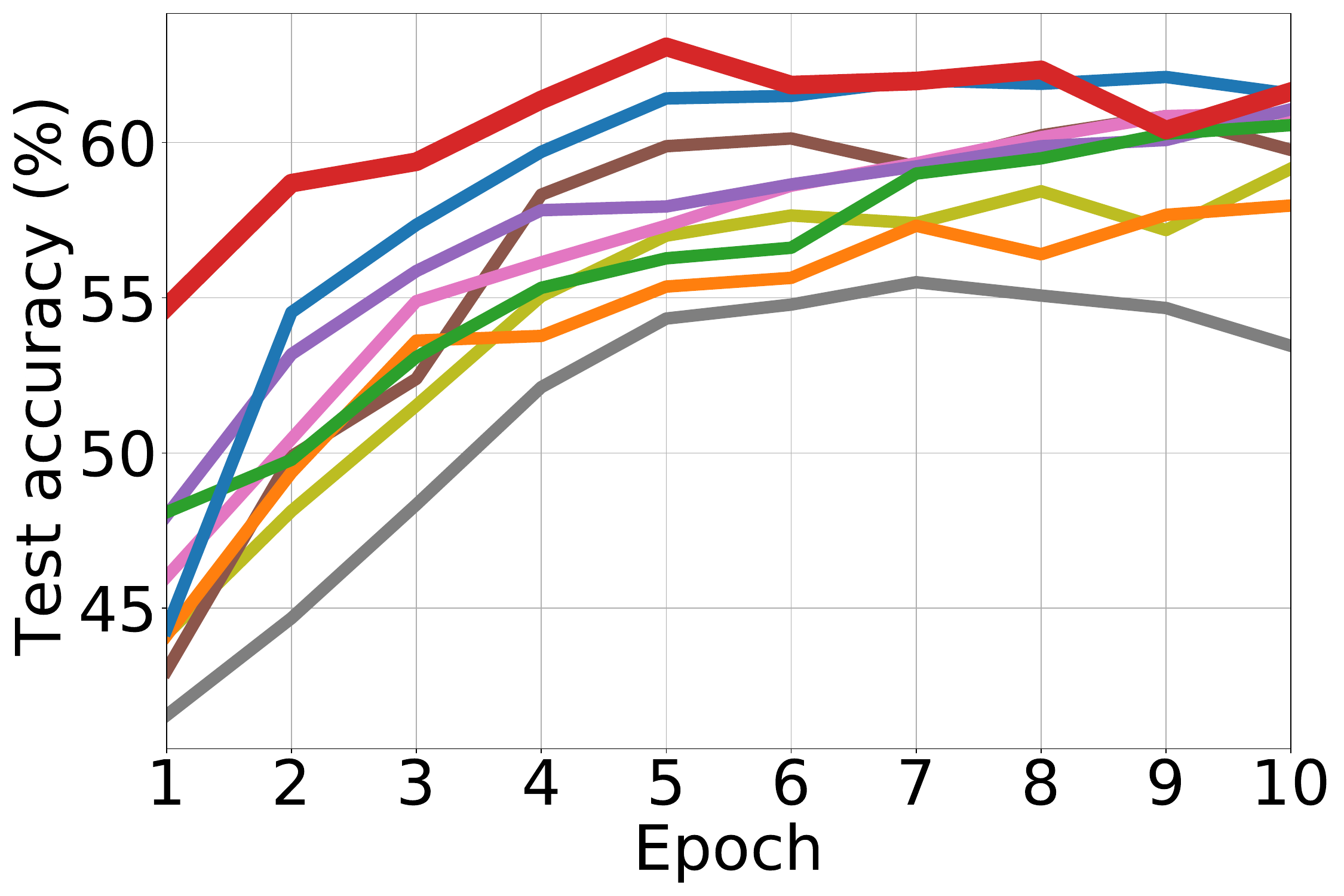}}
    \subfigure[][CIFAR-100]{\includegraphics[width=0.245\textwidth]{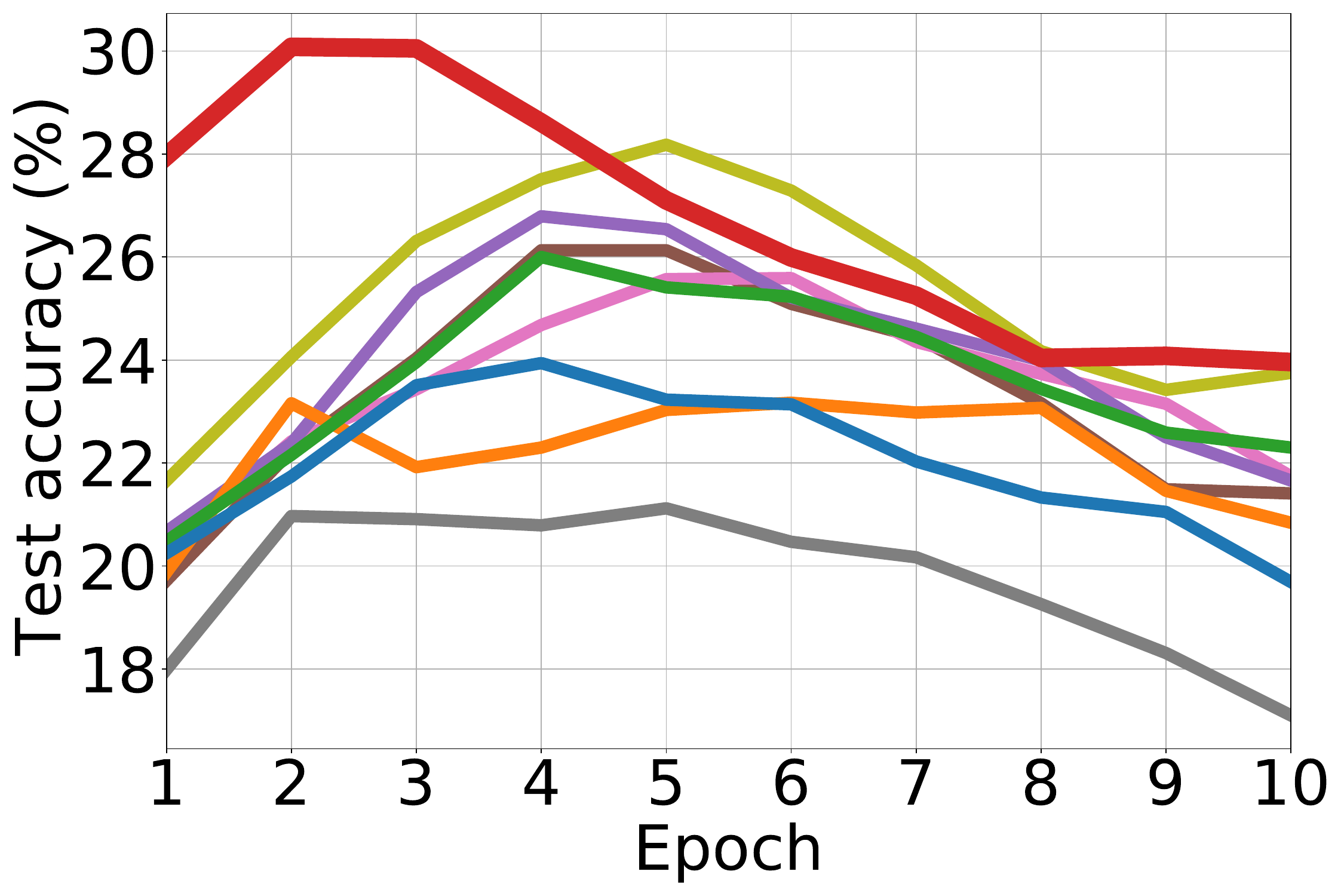}}
    \subfigure[][STL-10]{\includegraphics[width=0.245\textwidth]{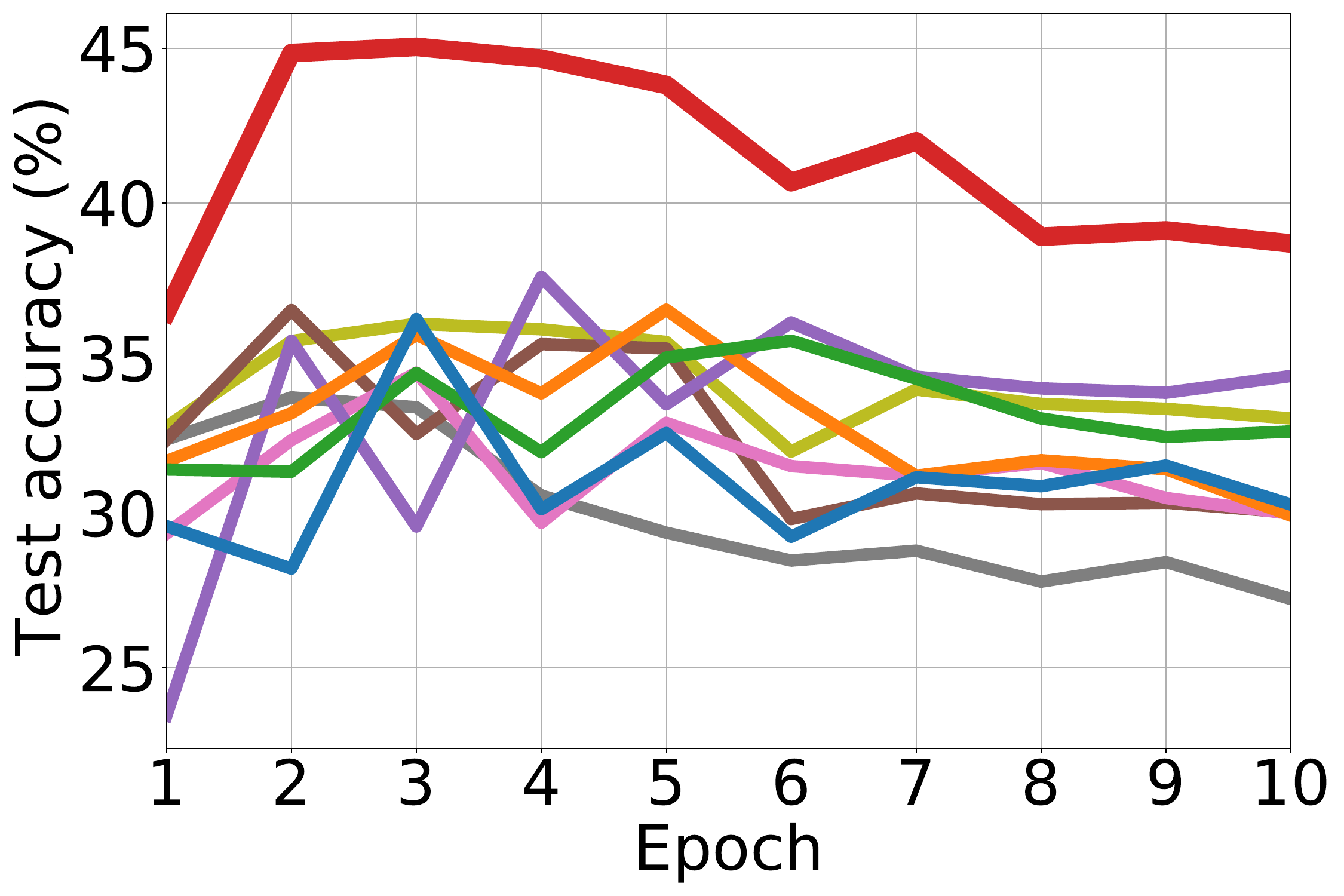}}
    \caption{The proposed BFNO-NODE outperforms existing state-of-the-art methods for test accuracy --- the results of train loss and NFE are in Appendix~\ref{sec:appendix_image}. We compare with other NODE-based models, such as vanilla NODE, ANODE, SONODE, HBNODE, GHBNODE, NesterovNODE, GNesterovNODE, and AdamNODE.}
    % \vspace{-1em}
    \label{fig:image_classification}
\end{figure*}

\paragraph{Dynamic global convolution $\rho$:} We note that in our method, the dynamic global convolutional operation $\rho$ is defined by the following method:
% \begin{linenomath}
\begin{align}
    \rho(\mathcal{F}(\mathbf{g}_{k})) &= FC(\mathbf{O}_1, \mathbf{O}_2, \cdots, \mathbf{O}_L),\label{eq:dg1}\\
    \mathbf{O}_i &= \mathbf{R}_{i} {\odot} \mathcal{F}(\mathbf{g}_{k}),\label{eq:dg2}
% \mathbf{R}_i &= \sigma(FC_i(\mathcal{F}(\mathbf{g}_{k}))),
%\label{eq:r_opt}
\end{align}
% \end{linenomath}

where $FC$ means a fully-connected layer, and $\mathbf{O}_i$ is the elementwise multiplication between the $i$-th global convolutional kernel $\mathbf{R}_i$ and $\mathcal{F}(\mathbf{g}_k)$. Therefore, $FC$ is to learn how to dynamically aggregate the $L$ different global convolution outcomes, i.e., $\{\mathbf{O}_1, \mathbf{O}_2, \cdots, \mathbf{O}_L\}$, where $L$ is a hyperparameter. The learnable parameters (kernels) in $\mathbf{R}_i$, $1 \leq i \leq L$ and $FC$ constitute our proposed dynamic global convolutional operation with respect to the input $\mathcal{F}(\mathbf{g}_k)$.
% \paragraph{Training:} Our method does not require special training methods, but existing training methods of NODEs can  be used without any changes after defining an ODE function based on BFNO-NODE.

\section{Experiments}
% \section{Evaluation}
\label{sec:Experiments}

In this section, we compare our model with existing baseline models for three general machine learning tasks: image classification, time series classification, and image generation. In comparison with the experiment set of HBNODE~\cite{xia2021heavy}, our experiment set covers more general machine learning tasks and several more datasets. Our software and hardware environments are as follows: \textsc{Ubuntu} 18.04 LTS, \textsc{Python} 3.6, \textsc{TorchDiffEq}, \textsc{Pytorch} 1.10.2, \textsc{CUDA} 11.4, \textsc{NVIDIA} Driver 470.74, i9 CPU, and \textsc{NVIDIA RTX A6000}.

\subsection{Image classification}
\label{sec:image_classification}
\paragraph{Datasets:}We test baselines and our model with the following four image classification benchmarks: MNIST~\cite{lecun2010mnist}, CIFAR-10~\cite{CIFAR}, CIFAR-100~\cite{CIFAR}, and STL-10~\cite{coates2011analysis}.

\paragraph{Experimental environments:}In this experiment, NODE~\cite{chen2018neural}, ANODE~\cite{dupont2019augmented}, SONODE~\cite{norcliffe2020second}, HBNODE, GHBNODE~\cite{xia2021heavy}, NesterovNODE, GNestrovNODE~\cite{nguyenimproving}, and AdamNODE~\cite{cho2022adamnodes} are used as baselines, which cover a prominent set of enhancements for NODEs.

We use a learning rate of 0.001 and a batch size of 64. In all datasets, we use DOPRI with its default tolerance settings to solve the integral problem. For HBNODE and GHBNODE, the damping parameter $\gamma$ is set to $sigmoid(\pi)$, where $\pi$ is trainable and initialized to -3. For all NODE-based baselines, we use three convolutional layers to define their ODE functions and for our method, we use two BFNO layers ($N=2$) and two kernels ($L=2$). For fair comparison, their model sizes (i.e., the number of parameters) are almost the same. (cf. Table~\ref{tab:Imageclassification_param_acc}).

In order to evaluate in various aspects, we consider the following evaluation metrics: i) the convergence speed of train loss, and most importantly ii) test accuracy. In addition, the number of function evaluations (NFEs) is an effective metric to measure the complexity of the forward and backward computation for NODE-based models. We focus on the accuracy in the main paper and refer readers to Appendix~\ref{sec:appendix_image} for additional analyses on the NFEs and train loss curves.

\paragraph{Experimental results:} Figure~\ref{fig:image_classification} shows the test accuracy curve in each dataset, and Table~\ref{tab:Imageclassification_param_acc} summarizes the overall results. Our method, shown in red, had the fastest convergence speed, outperforming many baselines. For three datasets except for MNIST, our method shows far better test accuracy curves than those of other baselines. Other baselines' train loss values were much higher than those of our method in MNIST. %Our method also shows good results for the test accuracy.
In particular, our method dominates the test accuracy in STL-10, achieving higher accuracy consistently. For MNIST, however, AdamNODE shows a higher accuracy than ours.
Appendix~\ref{sec:appendix_image} contains the train loss and NFE curves. In general, our method's  forward and backward NFEs are the smallest (or close to the smallest), which is an outstanding achievement. Our method enhances not only the test accuracy but also the computational complexity. In Appendix~\ref{a:tsne}, we also visualize the feature maps from various methods. While vanilla NODEs' feature map shows some glitches, ours shows much better clustering outcomes.

\begin{table}[t]
\centering
	    \small
        \setlength{\tabcolsep}{4pt}
        \renewcommand{\arraystretch}{1.2}
        \begin{tabular}{lc}
        \specialrule{1pt}{1pt}{1pt}
        Method & Accuracy   \\ \hline
        RNN-$\Delta_{t}$ & 0.797 $\pm$ 0.003\\
        RNN-Impute &  0.795 $\pm$ 0.008 \\ 
        RNN-D &  0.800 $\pm$ 0.010 \\
        GRU-D      &  0.806 $\pm$ 0.007 \\ 
        RNN-VAE         &  0.343 $\pm$ 0.040  \\
        ODE-RNN & 0.829 $\pm$ 0.016 \\ 
        GHBNODE-RNN & 0.838 $\pm$ 0.017 \\
        GNesterovNODE-RNN & 0.840 $\pm$ 0.016 \\
        % HBNODE-RNN & 0.821 $\pm$ 0.015 \\
        Latent-ODE (RNN enc.) & 0.835 $\pm$ 0.010 \\
        Latent-ODE (ODE enc.) & 0.846 $\pm$ 0.013 \\
        Augmented-ODE & 0.811 $\pm$ 0.031 \\
        ACE-Latent-ODE & 0.859 $\pm$ 0.011 \\ \hline 
        % BFNO-NODE-RNN & {\color{red}{0.839 $\pm$ 0.006}}  \\ 
        Latent-BFNO-NODE (ODE enc.) &  \textbf{0.874 $\pm$ 0.004} \\
        % & FFJORD ORIGIN & CIFAR-10  \\ \hline
        % MNIST       & 0.99 & 3.40             \\ 
        % CIFAR-10           & 0.97 & 3.38    \\   \hline
        % FFJORD CWNODE (OURS)   &\textbf{0.88}         &      \textbf{3.33}               \\
        \specialrule{1pt}{1pt}{1pt}
        \end{tabular}
        \caption{Time series classification (HumanAct.)}
        \label{tab:human_activity}
\end{table}

\subsection{Time series classification}

% models are trained and evaluated with benchmark datasets Physionet and HumanActivity.
\paragraph{Datasets:}HumanActivity~\cite{kaluvza2010activity} and Physionet~\cite{doi:10.1177/1460458219850323} benchmark datasets are used to train and evaluate models for time series classification. The HumanActivity dataset includes information from five people with four sensors at their left ankle, right ankle, belt, and chest while performing a variety of activities such as walking, falling, lying down, rising from a lying position, etc. Each activity of a person was recorded five times for the sake of reliability. In this experiment, we classify each person's input into one of the seven activities. 

PhysioNet consists of 8,000 time series samples and is used to forecast the mortality of intensive care unit (ICU) populations. The dataset had been compiled from 12,000 ICU stays. They documented up to 42 variables and removed brief stays of less then 48 hours. Additionally, the data recorded in this way have a timestamp of the elapsed time since admission to the ICU. In this task, the patient's life or death is determined based on records.

\paragraph{Experimental environments: } We consider a variety set of RNN-based models, ODE-RNN, and Latent-ODE, following the evaluation protocol in~\cite{rubanova2019latent}. In addition, we build two more baselines by replacing the ODE function of ODE-RNN with various methods: i) GHBNODE-based design to enhance ODE-RNN, denoted by GHBNODE-RNN; ii) GNesterov-based design, denoted by GNesterovNODE-RNN. To enhance Latent-ODE, we adopt the attention mechanism for NODEs proposed in~\cite{jhin2021ace} and call it ACE-Latent-ODE. Our model, Latent-BFNO-NODE, is implemented by replacing the ODE function of Latent-ODE (ODE enc.) with the BFNO-NODE-based design. In this process, the hyperparameter $L$ is fixed to 1 for efficiency.
We use accuracy for HumanActivity (since the dataset is balanced). AUROC is used for Physionet, considering its imbalanced nature.

%         GHBNODE-RNN & 0.838 $\pm$ 0.017 & 0.513 $\pm$ 0.001\\
%         GNesterovNODE-RNN & 0.840 $\pm$ 0.016 & -\\

\begin{table}[t]
\centering
        \small
        \setlength{\tabcolsep}{4pt}
        \renewcommand{\arraystretch}{1.2}
        \begin{tabular}{lc}
        \specialrule{1pt}{1pt}{1pt}
        Method & AUROC   \\ \hline
        RNN-$\Delta_{t}$ & 0.787 $\pm$ 0.014\\
        RNN-Impute &  0.764 $\pm$ 0.016 \\ 
        RNN-D &  0.807 $\pm$ 0.003 \\
        GRU-D      &  0.818 $\pm$ 0.008              \\ 
        RNN-VAE         &  0.515 $\pm$ 0.040   \\  
        ODE-RNN & 0.833 $\pm$ 0.009 \\ 
        % GHBNODE-RNN & {\color{red}{0.513 $\pm$ 0.001}}\\
        % GNesterovNODE-RNN & -\\

        Latent-ODE (RNN enc.) & 0.781 $\pm$ 0.018\\
        % Latent ODE + Poisson & 0.826 $\pm$ 0.007\\ 
        Latent-ODE (ODE enc.) & 0.829 $\pm$ 0.004\\
        Augmented-ODE (ODE enc.) & 0.851 $\pm$ 0.002\\
        ACE-Latent-ODE (ODE enc.) & \textbf{0.853 $\pm$ 0.003} \\ \hline
        % Augmented ODE & 0.851 $\pm$ 0.002 \\ \hline
        % ACE-Latent ODE  & 0.853 $\pm$ 0.003 \\ \hline
        % BFNO-NODE-RNN & {\color{red}{0.827 $\pm$ 0.004}} \\
        Latent-BFNO-NODE (ODE enc.) & 0.852 $\pm$ 0.001\\

        % & FFJORD ORIGIN & CIFAR-10  \\ \hline
        % MNIST       & 0.99 & 3.40             \\ 
        % CIFAR-10           & 0.97 & 3.38    \\   \hline
        % FFJORD CWNODE (OURS)   &\textbf{0.88}         &      \textbf{3.33}               \\
        \specialrule{1pt}{1pt}{1pt}
        \end{tabular}
        \caption{Time series classification (PhysioNet)}
        \label{tab:physionet}
\end{table}

\paragraph{Experimental results:}
In Table~\ref{tab:human_activity}, we summarize the results for HumanActivity. In general, RNN-based models do not show good accuracy. RNN-VAE shows the worst accuracy. NODE-based models significantly outperform them. Our proposed Latent-BFNO-NODE shows the best accuracy, followed by ACE-Latent-ODE. Our method also marks a small standard deviation in accuracy, which shows the robust nature of our method.

Table~\ref{tab:physionet} summarizes the results for PhysioNet. ACE-Latent-ODE shows good performance for this dataset, which shows the appropriateness of its attention mechanism. However, our Latent-BFNO-NODE has an AUROC score comparable to it with a much smaller standard deviation.

\subsection{Image generation}
\paragraph{Datasets: } For our image generation task, we use MNIST and CIFAR-10. These two datasets are the most widely used for conducting generative task experiments for NODE-based and invertible neural network-based models~\cite{grathwohl2018ffjord}.

\paragraph{Experimental environments: }To verify the performance of the BFNO as ODE function, we consider the baselines in~\cite{finlay2020train}, which cover a prominent set of NODE-based and invertible neural network-based models. The ODE function $f(\mathbf{h}(t), t; \boldsymbol{\theta}_{f})$ in these baselines consists of four convolutional layers with $3\times3$ kernels and softplus activations. These layers contain 64 hidden units, and the time $t$ is concatenated to the spatial input vector $\mathbf{h}(t)$ (as a side channel). The Gaussian Monte-Carlo trace estimator is used to calculate the log-probability of a generated sample via the change of variable theorem.
% divergence of $f$, with one sample of fixed noise each solver time-step.
Using the Adam optimizer with a learning rate of 0.001, we train on a single GPU with a batch size of 200 for 100 epochs. We use a model with the kinetic regularization proposed in~\cite{finlay2020train}, and we propose FFJORD-BFNO-RNODE by replacing ODE function $f$ of FFJORD-RNODE with BFNO structure. Our model consists of three BFNO layers and the softplus activations in between them. The hyperparameter $L$ is fixed to 3. In addition, the transformation $\mathbf{W}$ of BFNO is replaced with a convolutional operation with a 3$\times$3 kernel since this task is generation.

\begin{table}[t]
\centering
\setlength{\tabcolsep}{8pt}
\scriptsize
\setlength{\tabcolsep}{5pt}
\renewcommand{\arraystretch}{1.2}
% \vspace{.5em}
\begin{tabular}{lcc}
\specialrule{1pt}{1pt}{1pt}
Method & MNIST & CIFAR-10  \\ \hline
REALNVP~\cite{dinh2016density} & 1.06 & 3.49 \\
I-RESNET~\cite{behrmann2019invertible} & 1.05 & 3.45\\ 
GLOW~\cite{kingma2018glow} & 1.05 & 3.35\\
FFJORD~\cite{grathwohl2018ffjord}       & 0.99 & 3.40             \\ 
FFJORD-RNODE~\cite{finlay2020train}           & 0.97 & 3.38  \\   \hline
FFJORD-BFNO-RNODE   &\textbf{0.88}         &      \textbf{3.33}           \\
% & FFJORD ORIGIN & CIFAR-10  \\ \hline
% MNIST       & 0.99 & 3.40             \\ 
% CIFAR-10           & 0.97 & 3.38    \\   \hline
% FFJORD CWNODE (OURS)   &\textbf{0.88}         &      \textbf{3.33}               \\
\specialrule{1pt}{1pt}{1pt}
\end{tabular}
\caption{Negative log-likelihood (in bits/dim) on test images. Lower is better.}
\label{tab:ffjord}
\end{table}

\begin{figure}[t]
\centering
\subfigure[][Real images]{{\includegraphics[width=0.495\columnwidth ]{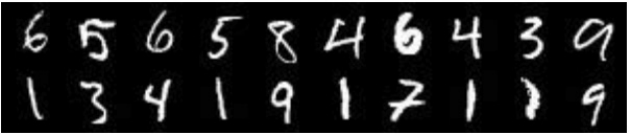} }}% 
\subfigure[][Real images]{{\includegraphics[width=0.495\columnwidth ]{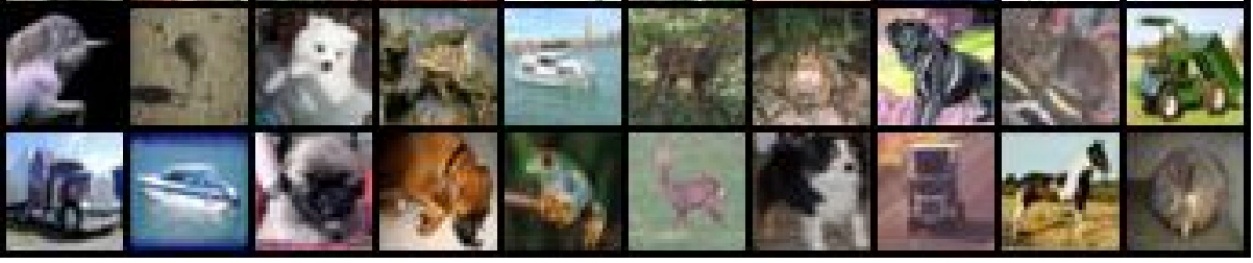} }}% 
\hfill 
\subfigure[][FFJORD-RNODE]{{\includegraphics[width=0.495\columnwidth ]{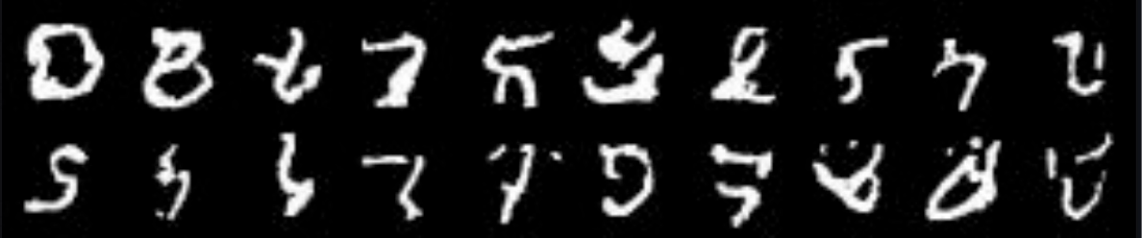} }}% 
\subfigure[][FFJORD-RNODE]{{\includegraphics[width=0.495\columnwidth ]{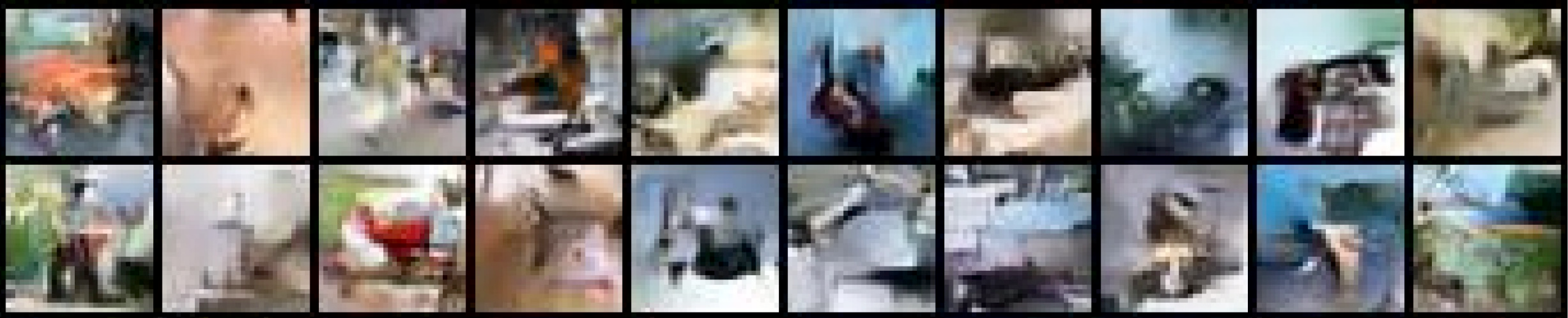} }}% 
\hfill 
\subfigure[][FFJORD-BFNO-RNODE]{{\includegraphics[width=0.495\columnwidth ]{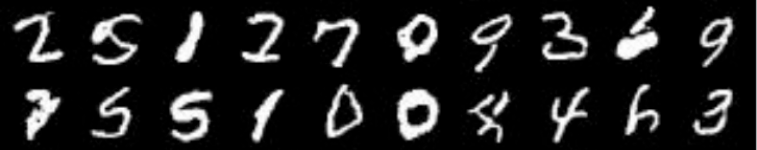} }}% 
\subfigure[][FFJORD-BFNO-RNODE]{{\includegraphics[width=0.495\columnwidth ]{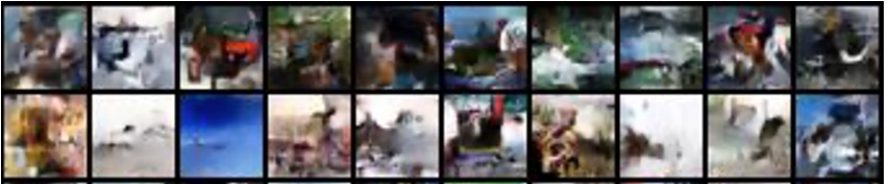} }}% 
\caption{Real and produced samples from models for MNIST (left) and CIFAR-10 (right). }%
\vspace{-1.2em}
\label{fig:ffjord}% 
\end{figure}
% \vspace{-1em}

\paragraph{Experimental results:} In Table~\ref{tab:ffjord}, it can be seen that the negative log-likelihood (NLL) of our model is better than those of all other baselines by large margins. In Figure~\ref{fig:ffjord}, in addition, we show real images and fake images by our method and FFJORD-RNODE, which has the best performance among the baselines. In general, all methods show acceptable fake images.

\begin{figure}[h]
    \centering
    \subfigure[][Ablation study on the number of parallelized global convolutions ($L$) (STL-10)]{\includegraphics[width=0.23\textwidth]{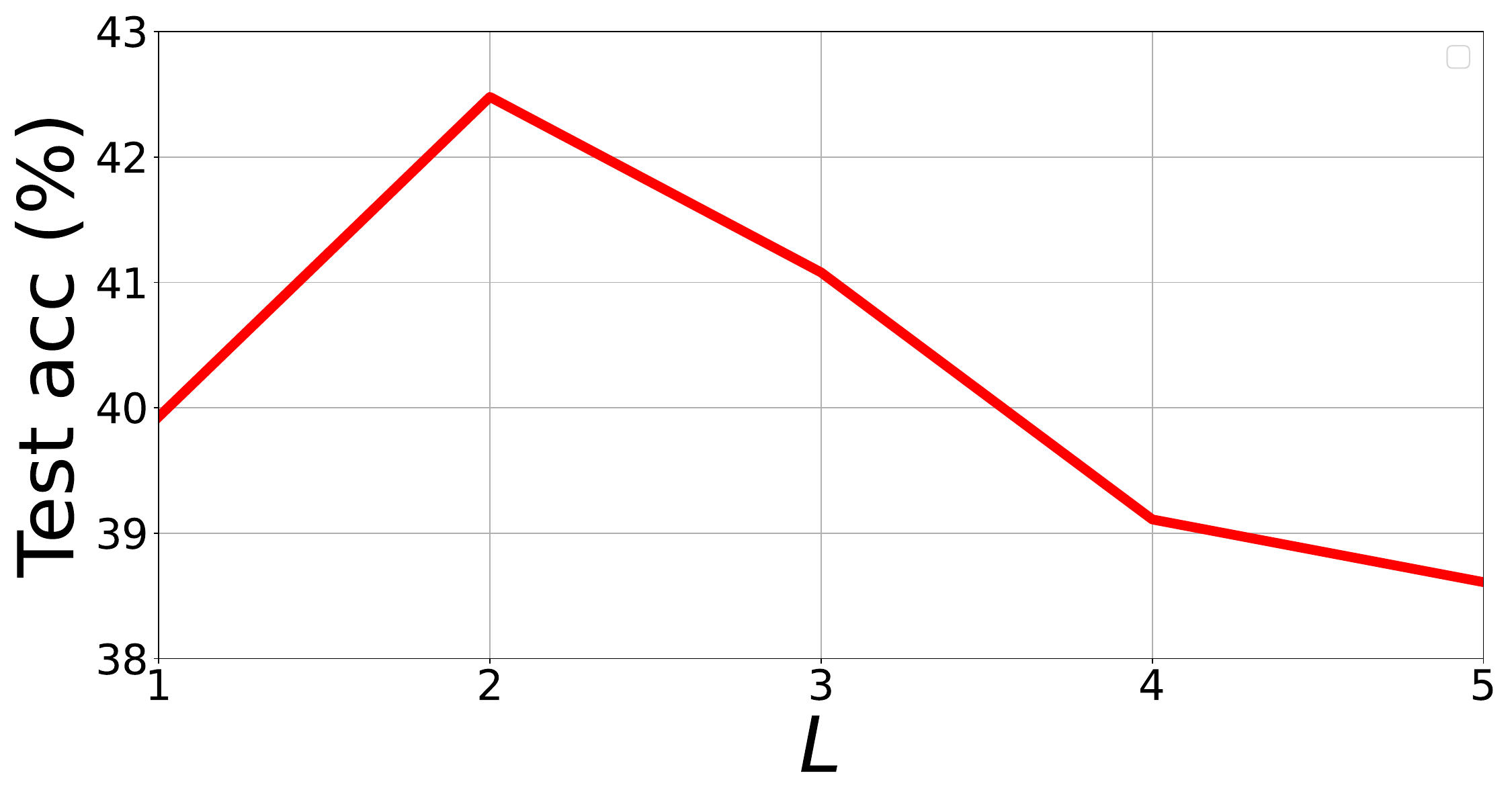}}\hfill
    \subfigure[][Ablation study on the neural operator type (STL-10)]{\includegraphics[width=0.23\textwidth]{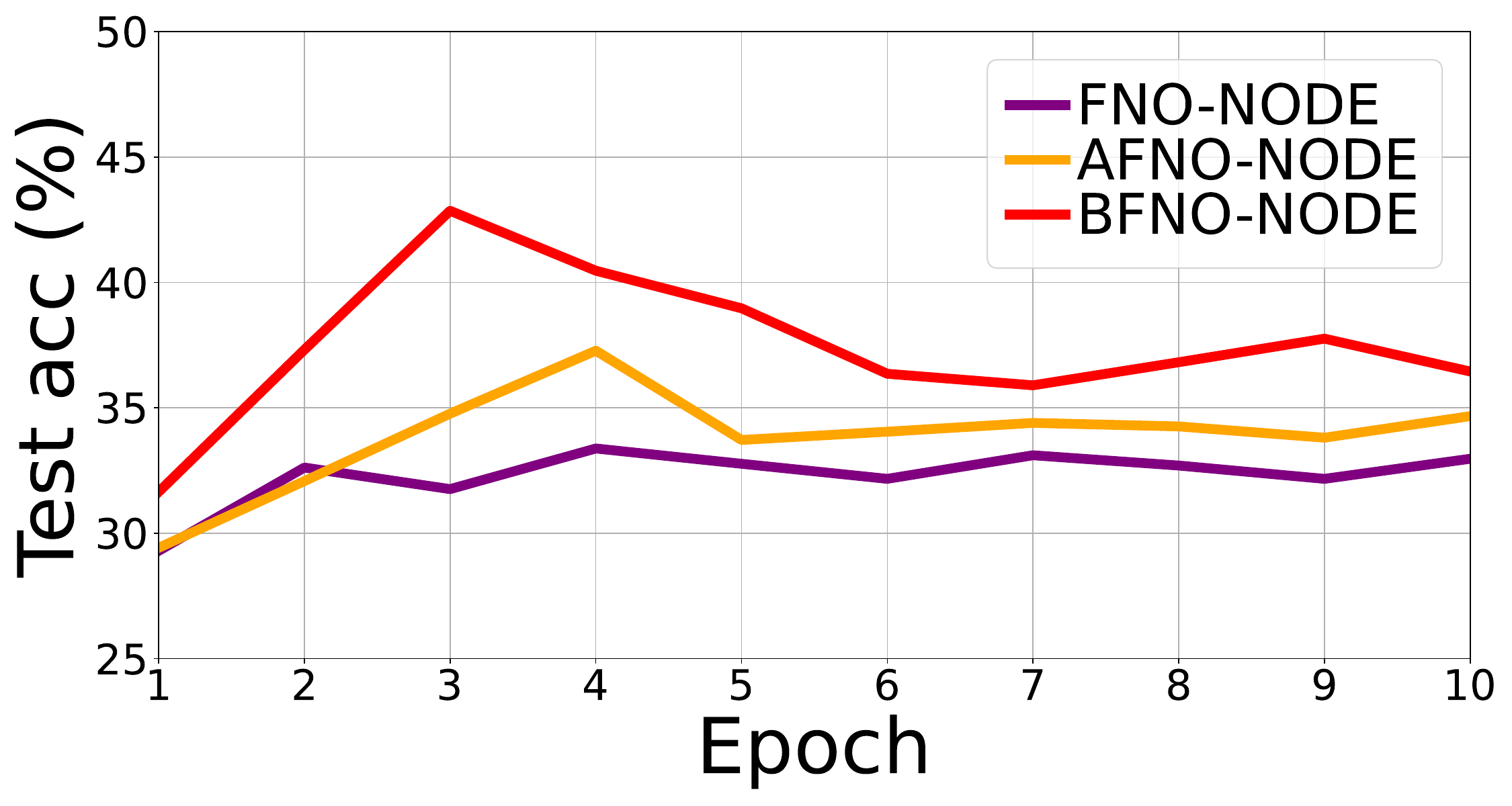}}
    % \subfigure[][Different resolution]{\includegraphics[width=0.32\textwidth]{images/ablation_2.pdf}}
    % \vspace{-.5em}
    \caption{Ablation study on the two key design options of BFNO. Other figures are in Appendix~\ref{sec:detail_ablation}.}
    % \vspace{-.5em}
    \label{fig:ablation_study}
    % \vspace{-1.em}
\end{figure}

\subsection{Ablation studies}
\label{ablation}
We conduct two key ablation studies: 1) checking the model accuracy by changing the number of parallelized global convolutions ($L$) in BFNO, and 2) comparing BFNO with existing neural operator methods, such as FNO and AFNO.
\paragraph{Accuracy by the number of parallelized global convolutions ($L$) in BFNO:} As shown in Figure~\ref{fig:detail_architecture} (b), a BFNO layer has multiple parallelized convolutions which are later merged by a fully-connected layer. The number of parallelized global convolutions in BFNO, denoted by $L$, is a key hyperparameter. Figure~\ref{fig:ablation_study} (a) shows the model accuracy for various settings for $L$ on image classification with STL-10. Too many or small convolutions lead to sub-optimal outcomes and $L=2$ is the best configuration in this ablation study. We fathom that using too many convolutions result in overfitting, which degrades the model accuracy.

\paragraph{Comparison with FNO and AFNO: } 
Figure~\ref{fig:ablation_study} (b) compares our method with FNO and AFNO. As shown, our BFNO-NODE consistently outperforms them throughout the entire training epoch. FNO was developed to model PDE operators, and AFNO aims at improving the vision transformer by replacing its spatial mixer with a special neural operator. Since they were not designed for NODEs, however, they fail to show effectiveness in our experiments when used to define the ODE function of NODEs. Similar patterns are observed in other datasets, and the results are shown in Appendix~\ref{sec:detail_ablation}.

\section{Conclusions}
Enhancing NODEs by adopting advanced ODE function architectures has been one active research trend for the past couple of years. However, existing approaches did not pay attention to the fact that the ODE function learns a special differential operator. In this work, we presented how neural operators can be used to define the ODE function and learn the differential operator. However, the na\"ive adoption of existing neural operators, such as FNO and AFNO, does not enhance NODEs. To this end, we designed a special neural operator architecture, called branched Fourier neural operators (BFNOs). Our dynamic global convolutional method with multiple parallelized global convolutions significantly improves the efficacy of various NODE-based models for three general machine learning tasks: image classification, time series classification, and image generation. Since the role of the ODE function is, in fact, applying a differential operator to the input $\mathbf{h}(t)$, our operator-based approach naturally shows the best fit in our experiments. Our ablation study with FNO and AFNO clearly shows that our BFNO design is one of the most important factors in our experiments.

% \section{Ethical Statement}
% \paragraph{Broader impact:}
\paragraph{Ethical Statement}
PhysioNet contains much personal health information. However, it was released after removing 90\% of observation to protect the privacy of the patients who have contributed their information. Therefore, our work does not have any related ethical concerns.

\section*{Acknowledgments}

This work was supported by the Institute of Information \& Communications Technology Planning \& Evaluation (IITP) grant funded at the Korean government (MSIT) (No. 2020-0-01361, Artificial Intelligence Graduate School Program by Yonsei University, 10\%), and (No.2022-0-00857, Development of AI/databased financial/economic digital twin platform, 90\%).

% \paragraph{Limitations:} Despite significantly enhancing the efficacy of NODEs for various tasks, our approach sometimes increases NFEs, resulting in larger computation in comparison with existing approaches (see Appendix~\ref{sec:appendix_image}). It is not straightforward to achieve both the high efficacy and the low computation at the same time. However, we believe that there exists an operator-based ODE function definition which simultaneously enhances the efficacy and the efficiency of NODEs.

\bibliography{aaai24}
\onecolumn

\newpage
\appendix
\onecolumn

\section{Various evaluation metrics in Image Classification}
\label{sec:appendix_image}

% {\color{red}{In general, our BFNO-NODE's NFE values are much smaller to those of other baselines. Our method also shows fast convergence speed for training loss.}}

\begin{figure}[h]
    \centering
    \includegraphics[width=0.95\textwidth]{images/figure/legend.pdf} %\vspace{-1em}

    \subfigure[][Train loss]{\includegraphics[width=0.245\columnwidth]{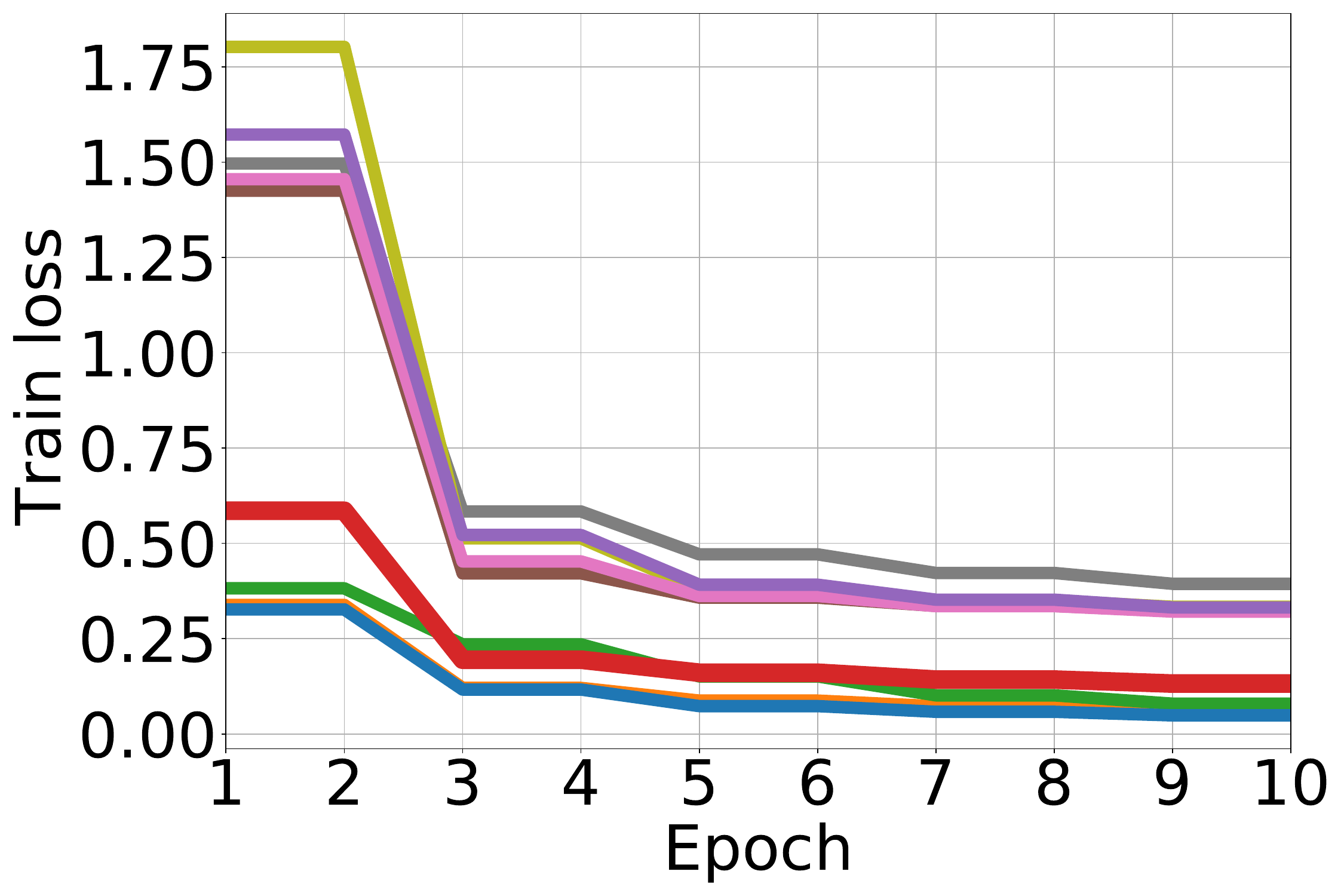}}
    \subfigure[][Train forward NFE]{\includegraphics[width=0.245\columnwidth]{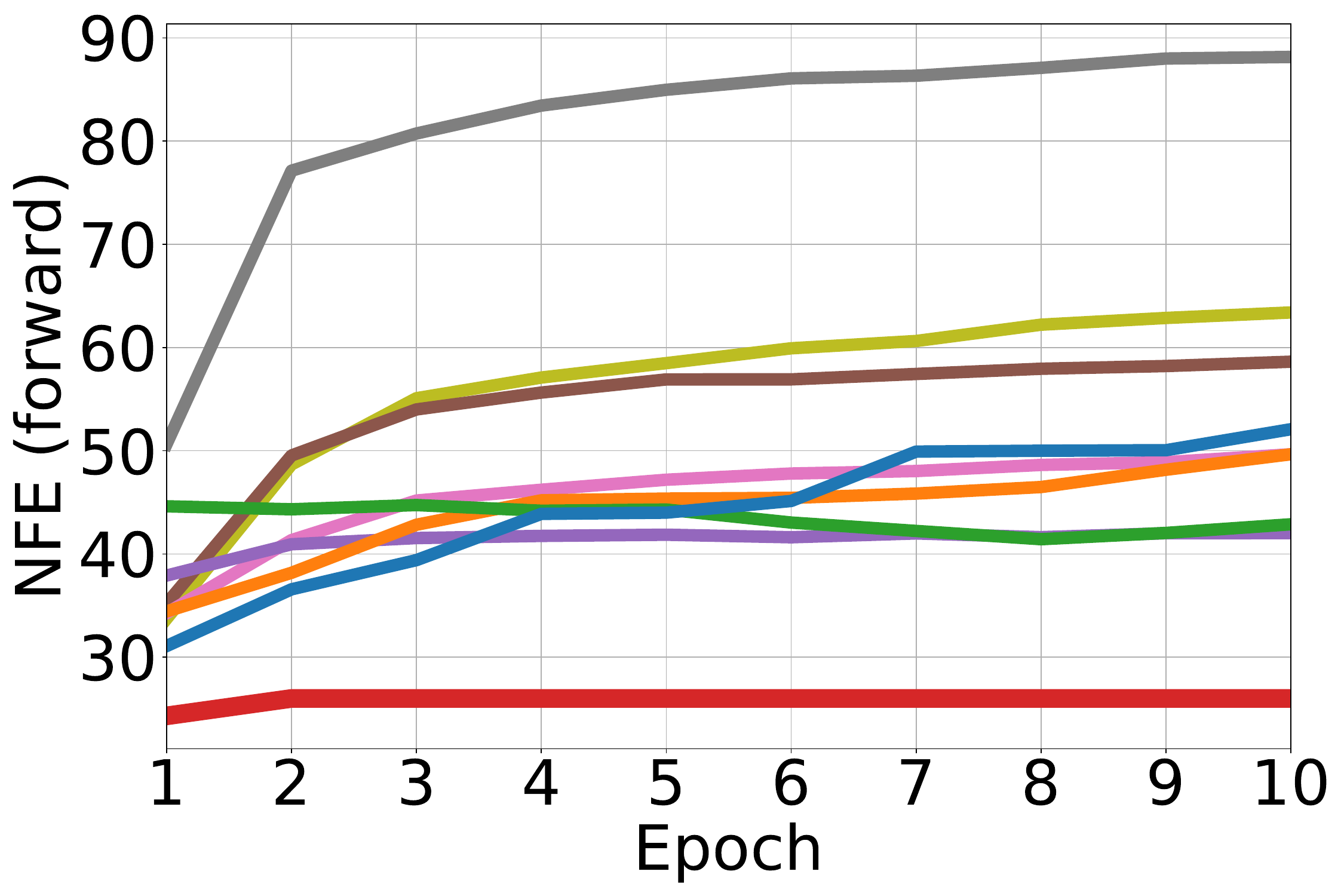}}
    \subfigure[][Train backward NFE]{\includegraphics[width=0.245\columnwidth]{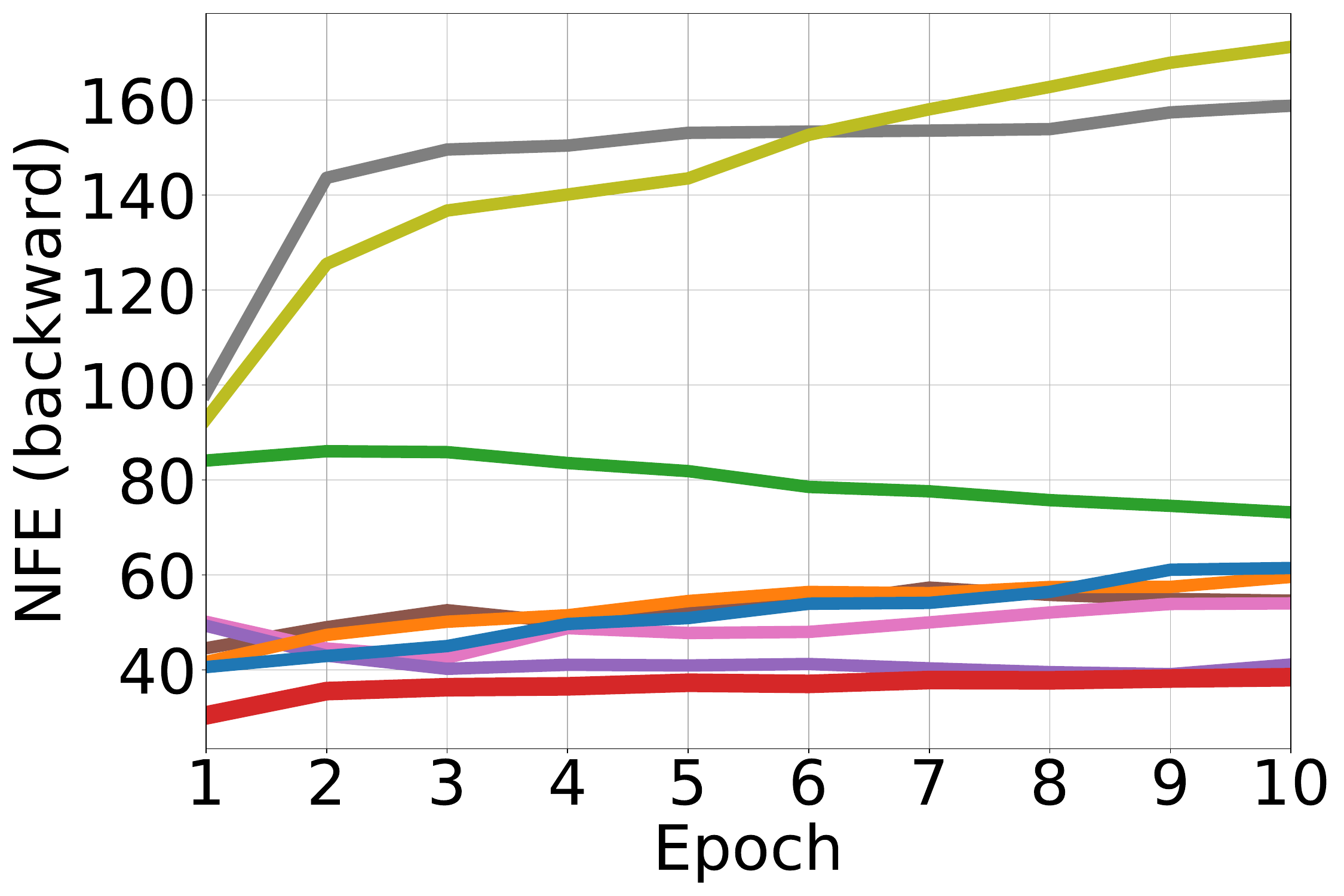}}
    \subfigure[][Test forward NFE]{\includegraphics[width=0.245\columnwidth]{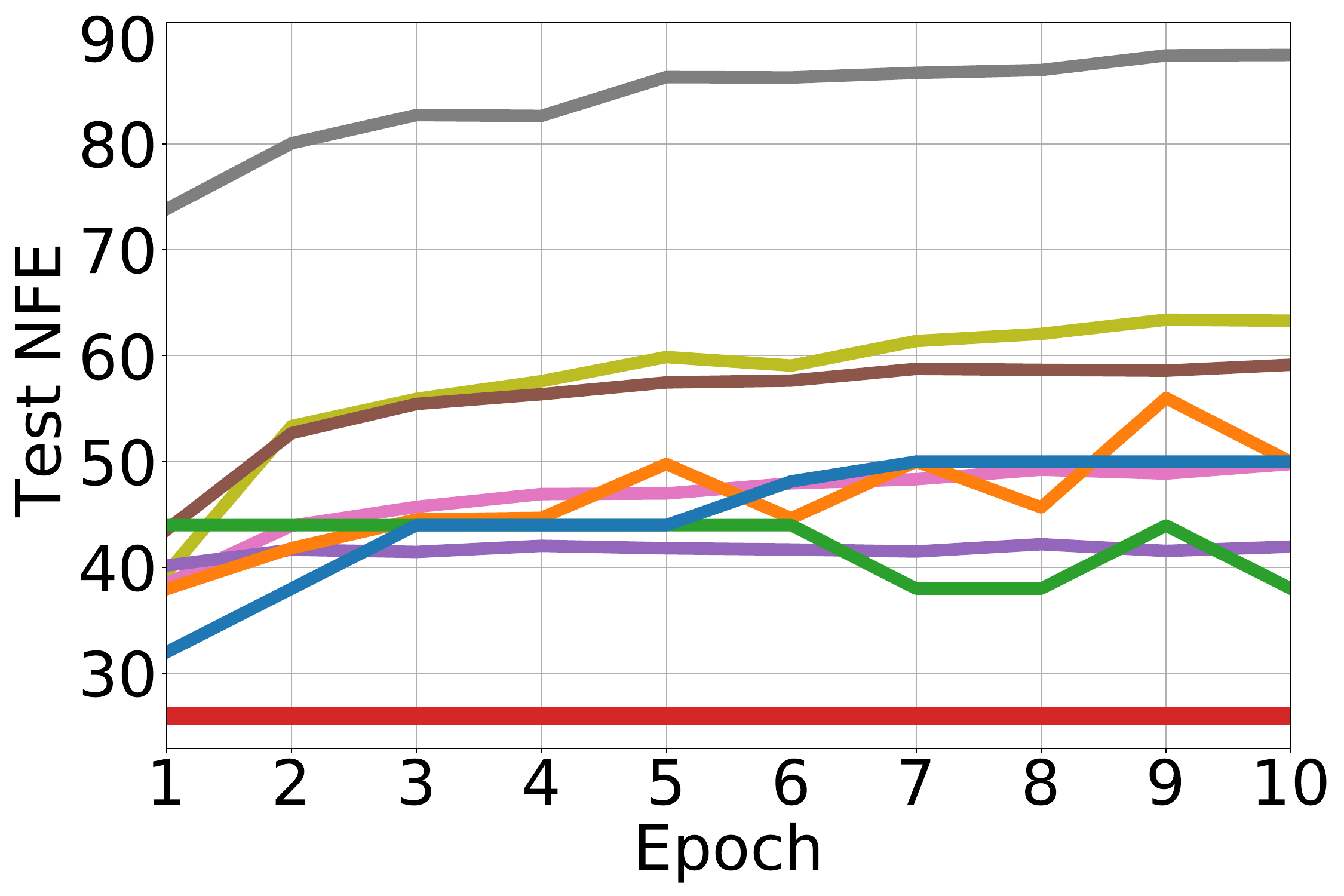}}
    
    \centering
    %
    % \vspace{-.5em}
    \caption{Train loss, train NFE (forward, backward), and test forward NFE on MNIST Classification}
    % \vspace{-1.3em}
    \label{fig:IC_MNIST}
\end{figure}

\begin{figure}[h]
    \centering
    \subfigure[][Train loss]{\includegraphics[width=0.245\columnwidth]{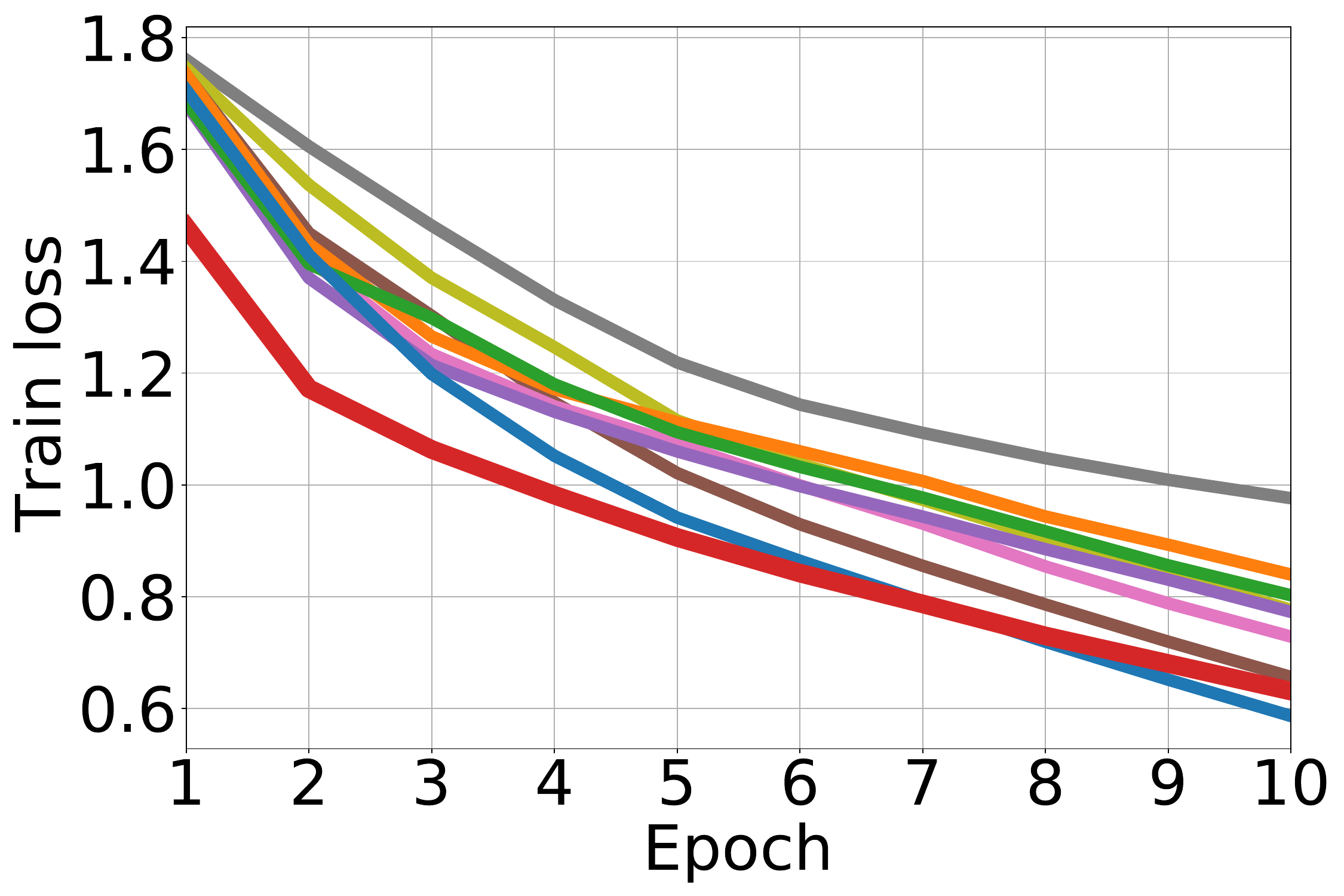}}
    \subfigure[][Train forward NFE]{\includegraphics[width=0.245\columnwidth]{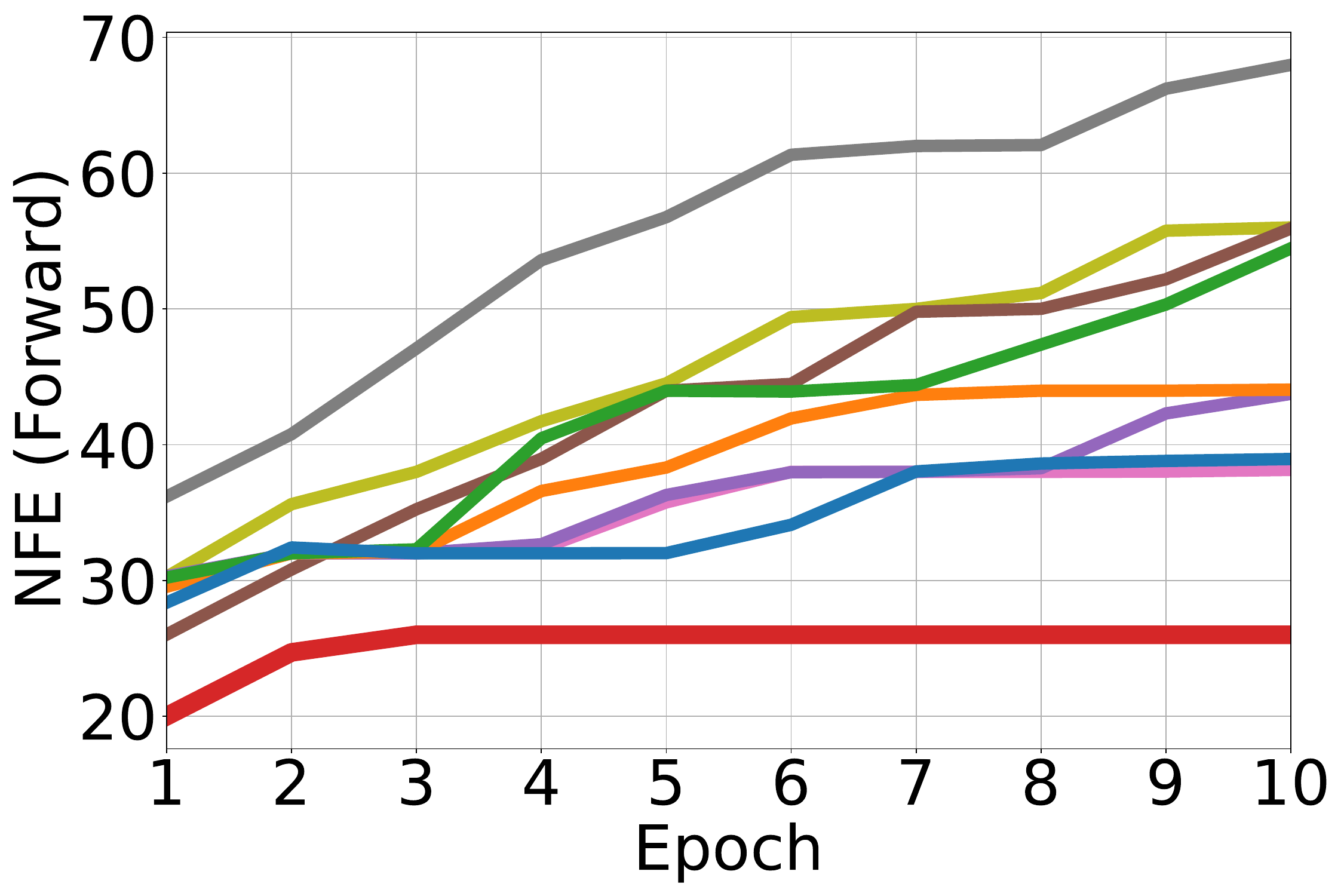}}
    \subfigure[][Train backward NFE]{\includegraphics[width=0.245\columnwidth]{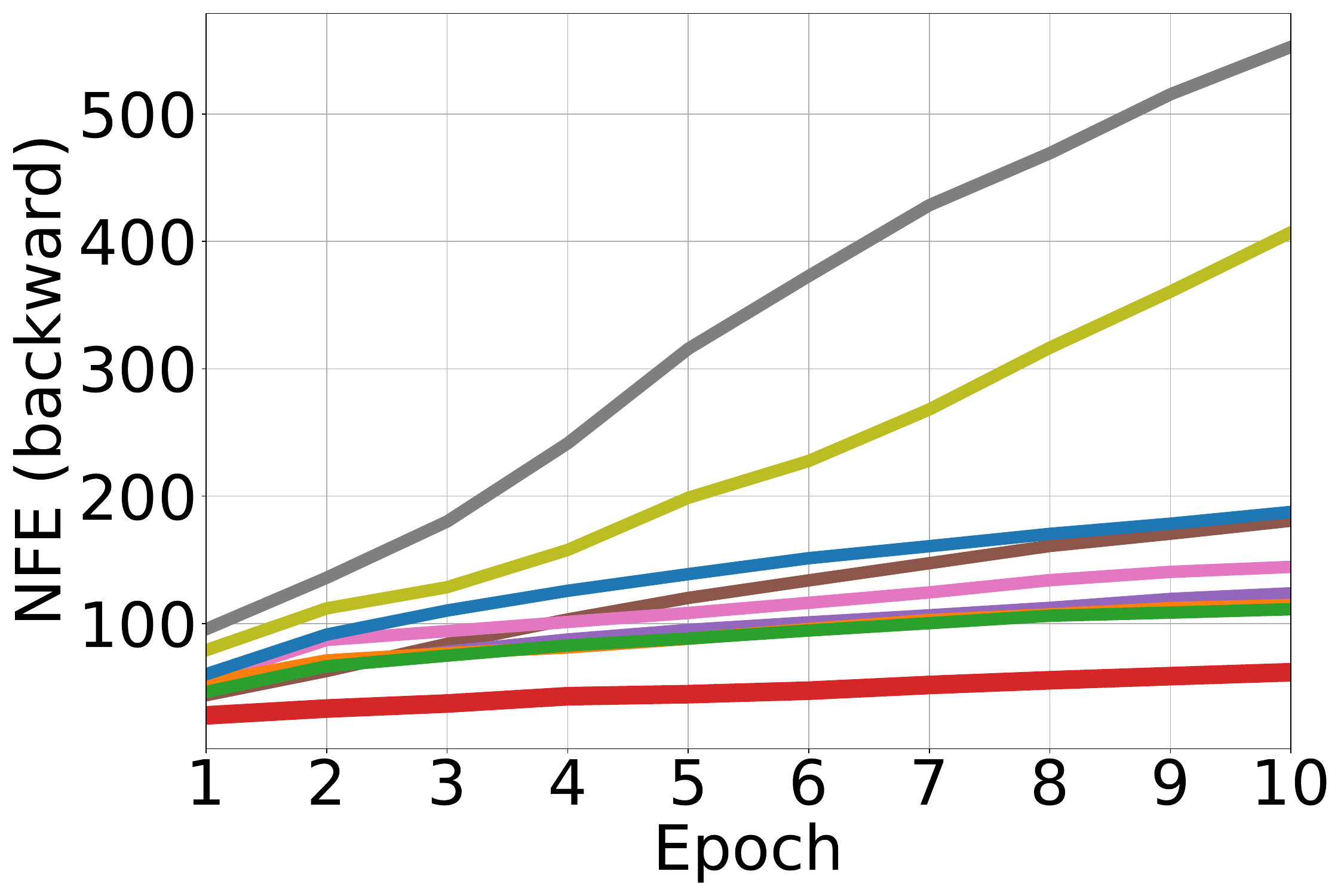}}
    \subfigure[][Test forward NFE]{\includegraphics[width=0.245\columnwidth]{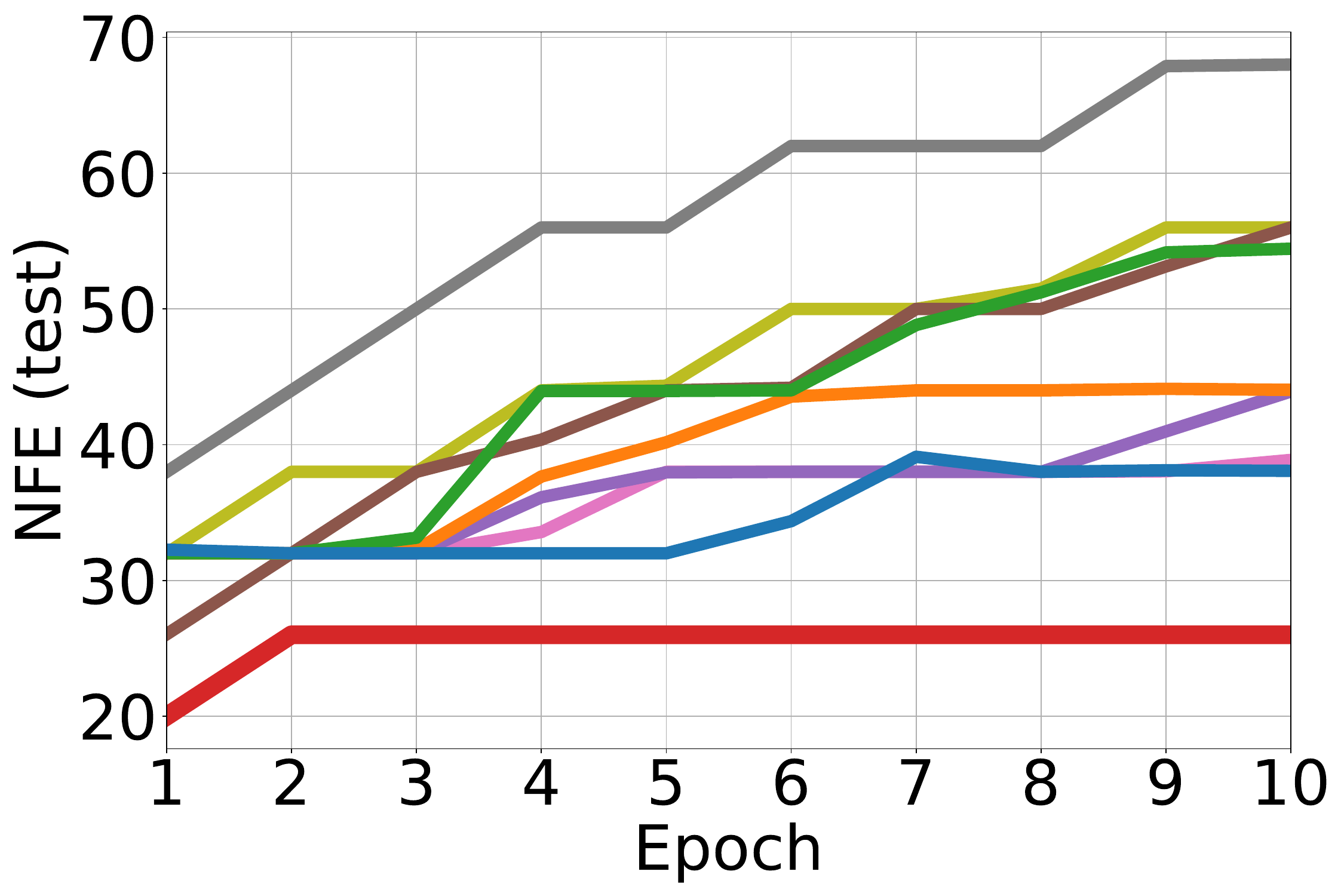}}
    
    \centering
    %
    % \vspace{-.5em}
    \caption{Train loss, train NFE (forward, backward), and test forward NFE on CIFAR-10 Classification}
    % \vspace{-1.3em}
    \label{fig:IC_CIFAR10}
\end{figure}

\begin{figure}[ht!]
    \centering
    \subfigure[][Train loss]{\includegraphics[width=0.245\columnwidth]{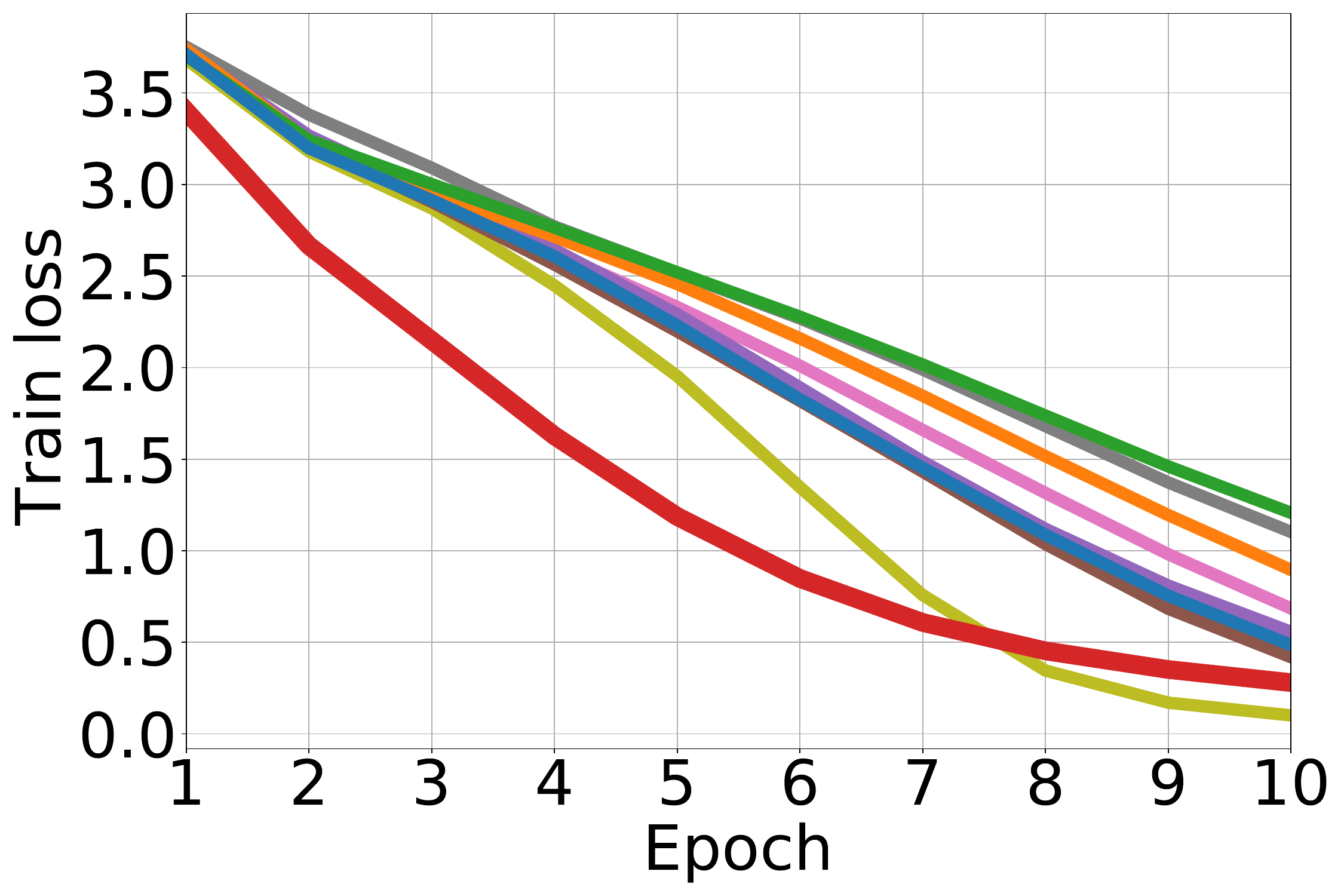}}
    \subfigure[][Train forward NFE]{\includegraphics[width=0.245\columnwidth]{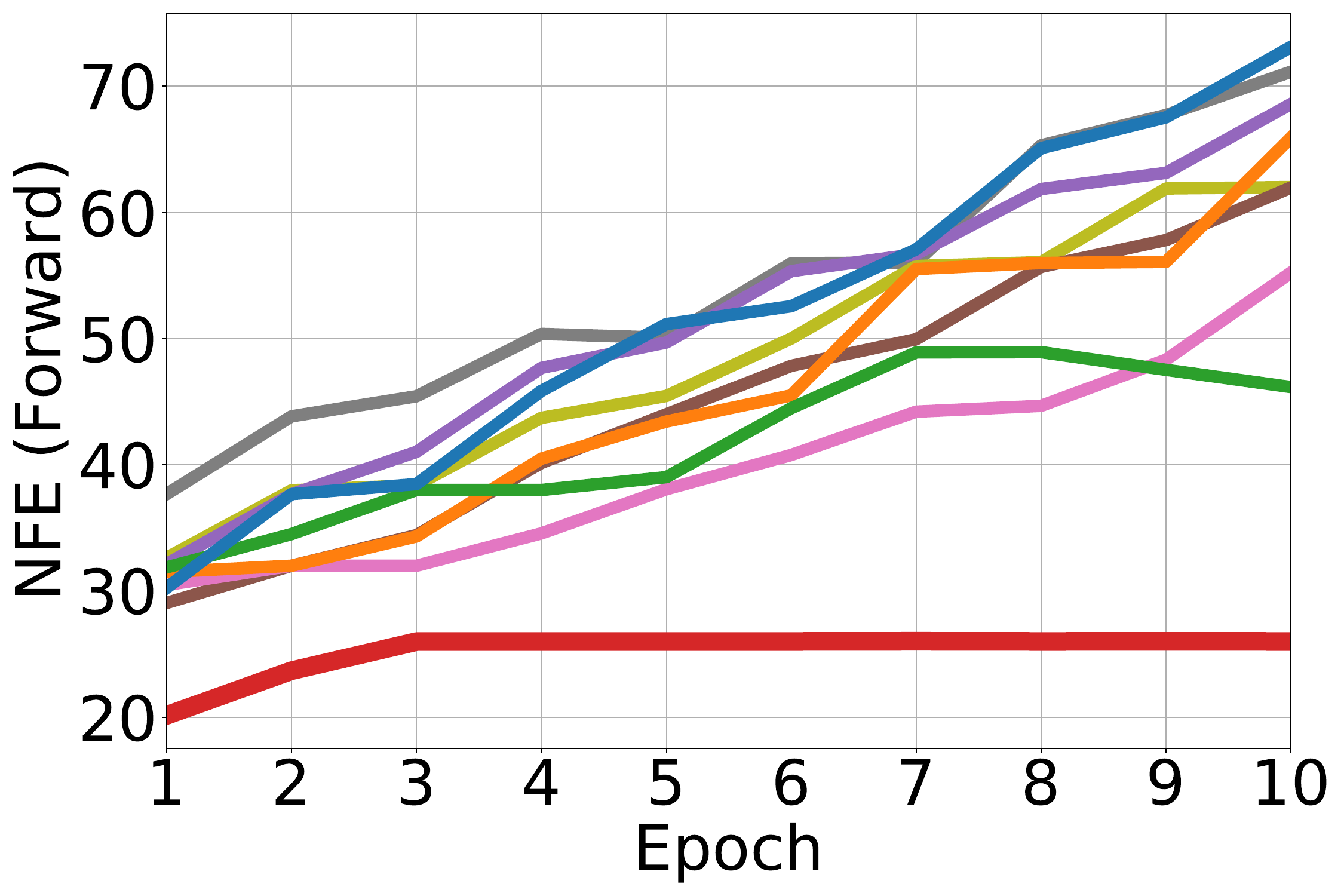}}
    \subfigure[][Train backward NFE]{\includegraphics[width=0.245\columnwidth]{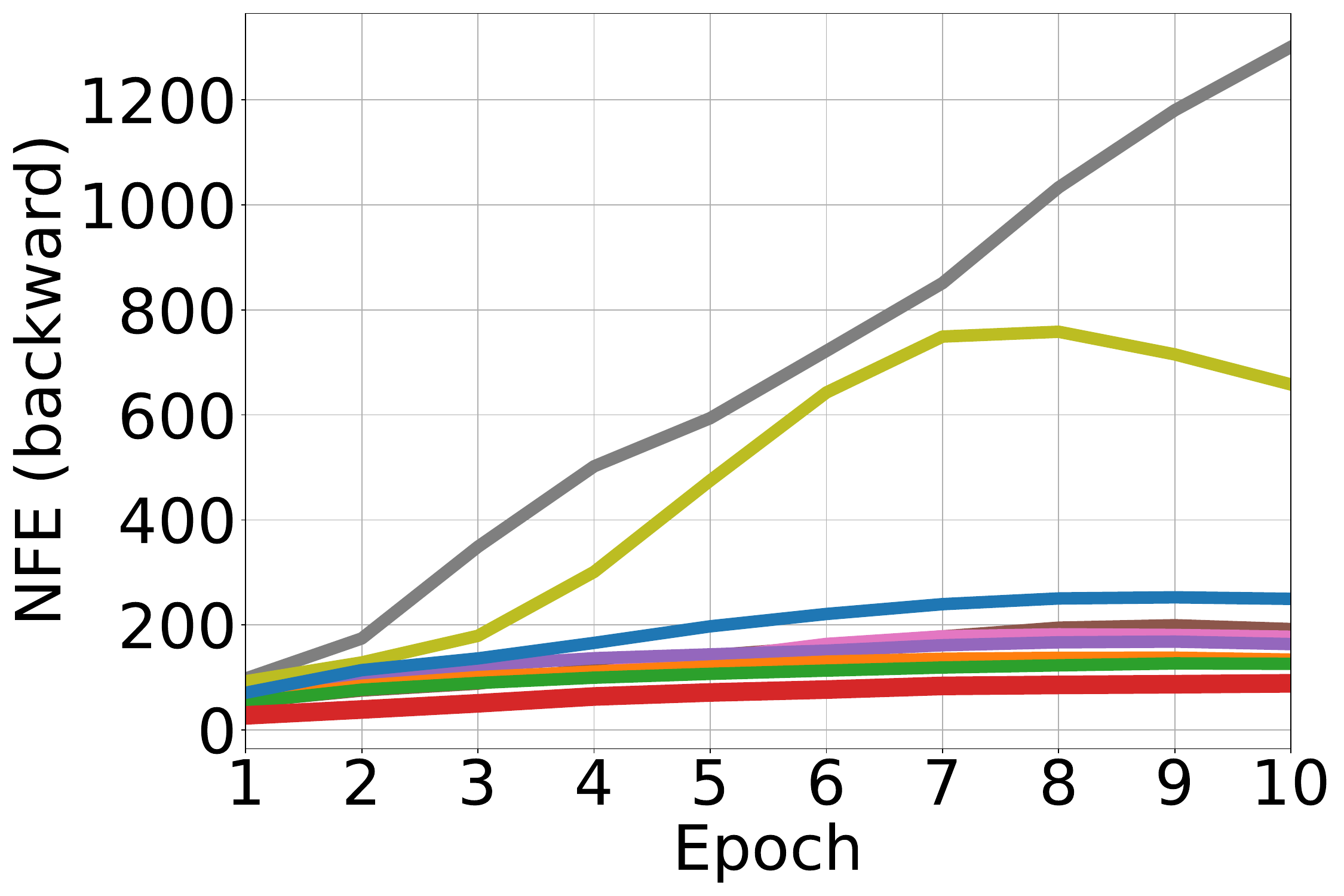}}
    \subfigure[][Test forward NFE]{\includegraphics[width=0.245\columnwidth]{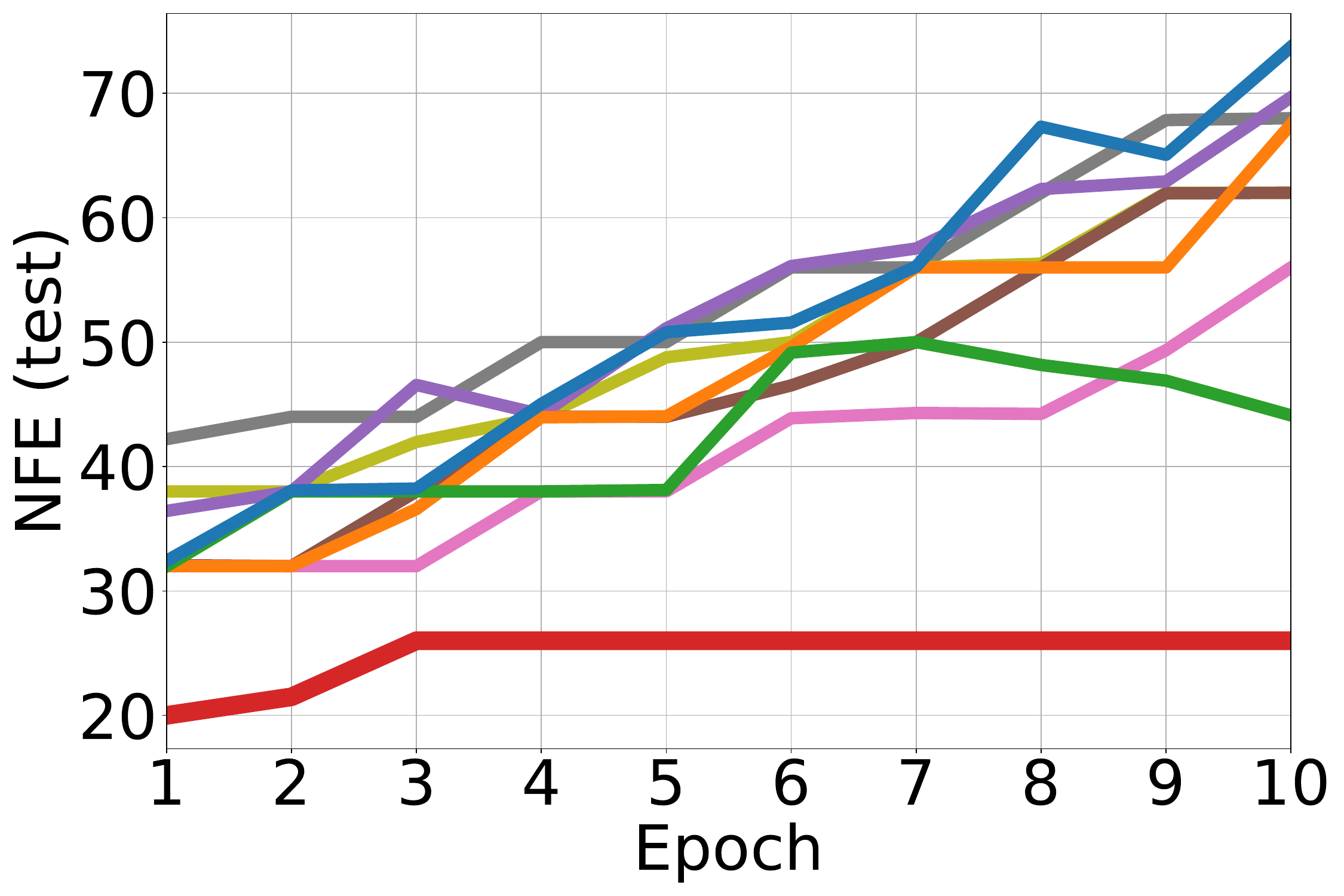}}
    
    \centering
    %
    % \vspace{-.5em}
    \caption{Train loss, train NFE (forward, backward), and test forward NFE on CIFAR-100 Classification}
    % \vspace{-1.3em}
    \label{fig:IC_CIFAR100}
\end{figure}

\begin{figure}[ht!]
    \centering
    \subfigure[][Train loss]{\includegraphics[width=0.245\columnwidth]{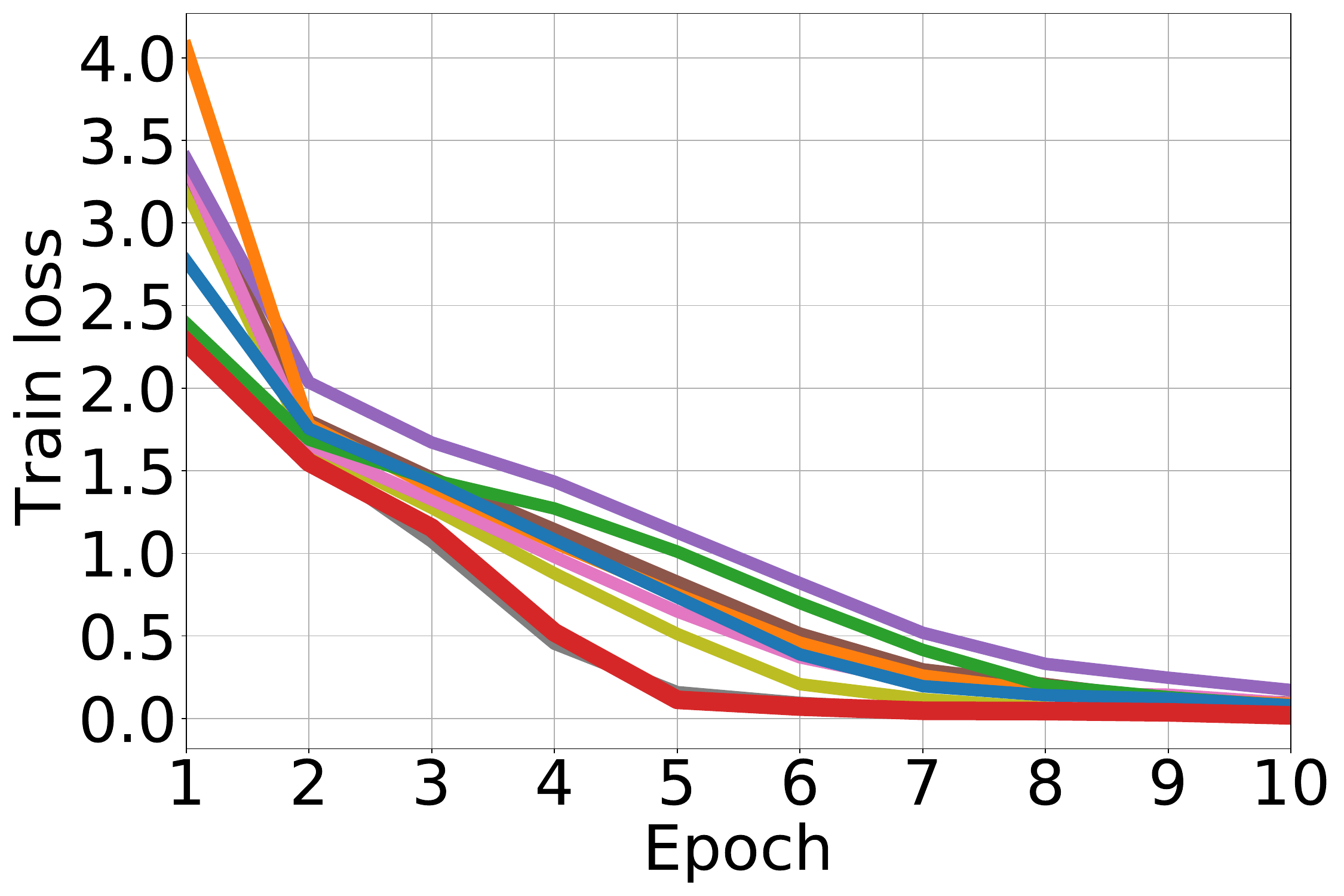}}
    \subfigure[][Train forward NFE]{\includegraphics[width=0.245\columnwidth]{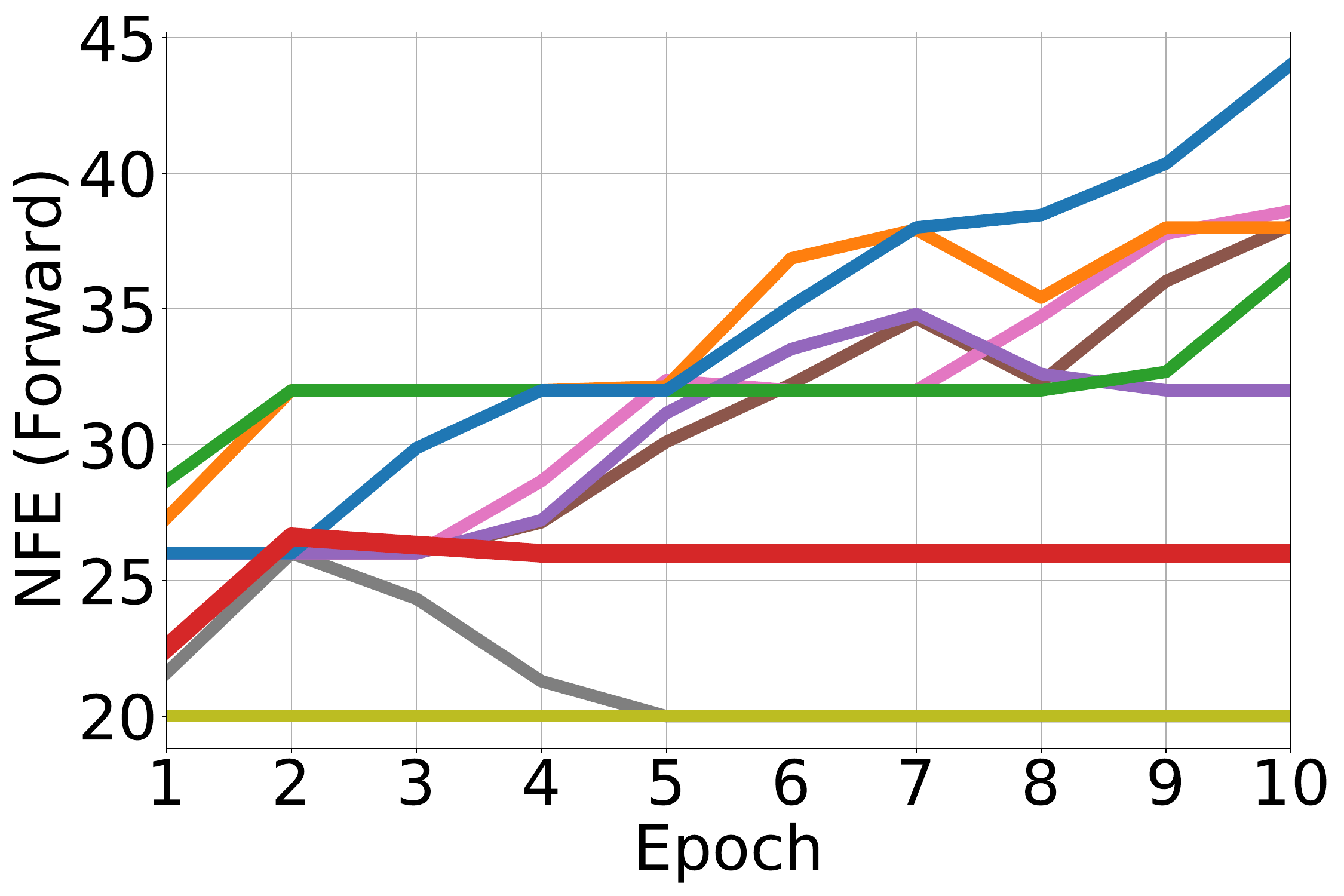}}
    \subfigure[][Train backward NFE]{\includegraphics[width=0.245\columnwidth]{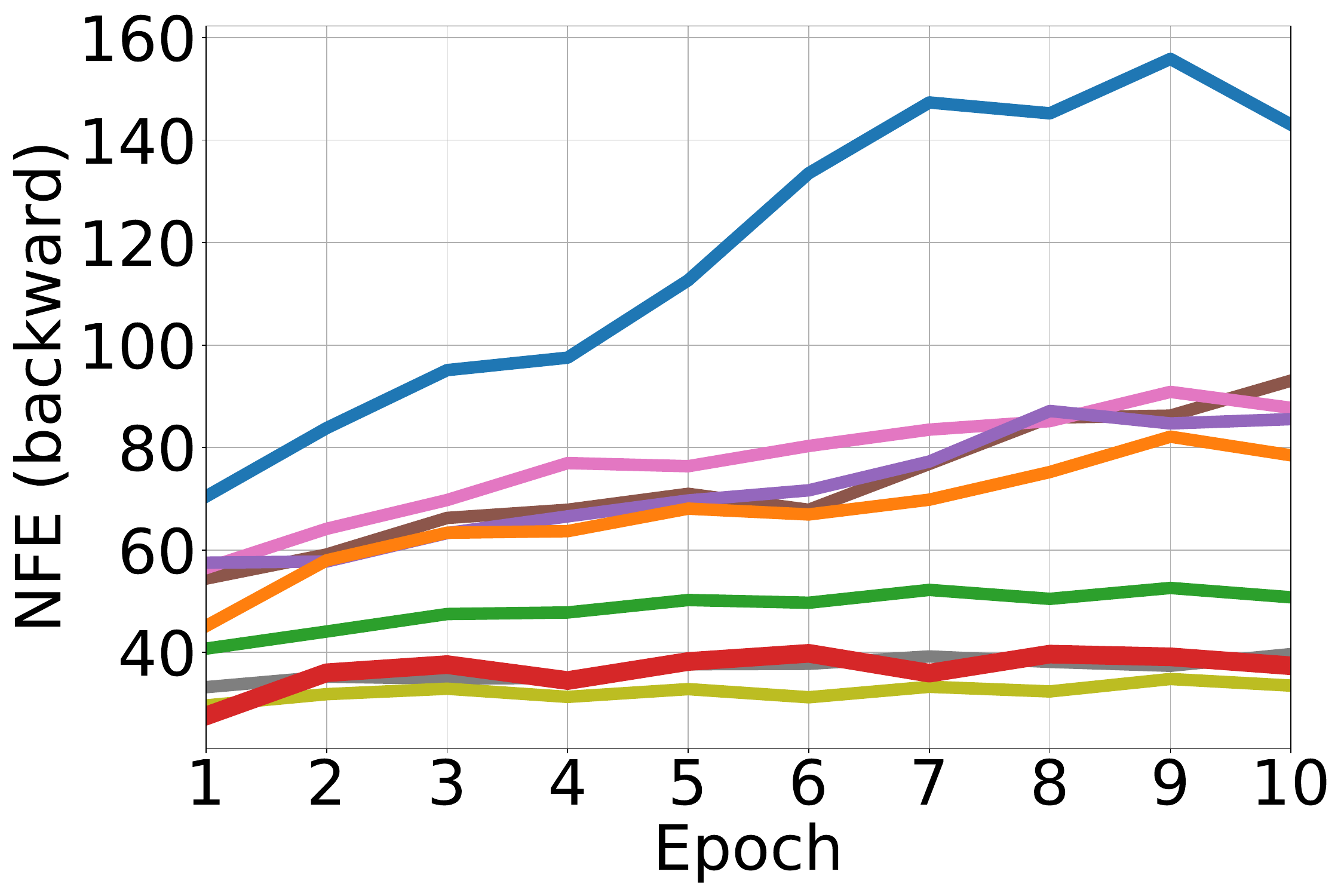}}
    \subfigure[][Test forward NFE]{\includegraphics[width=0.245\columnwidth]{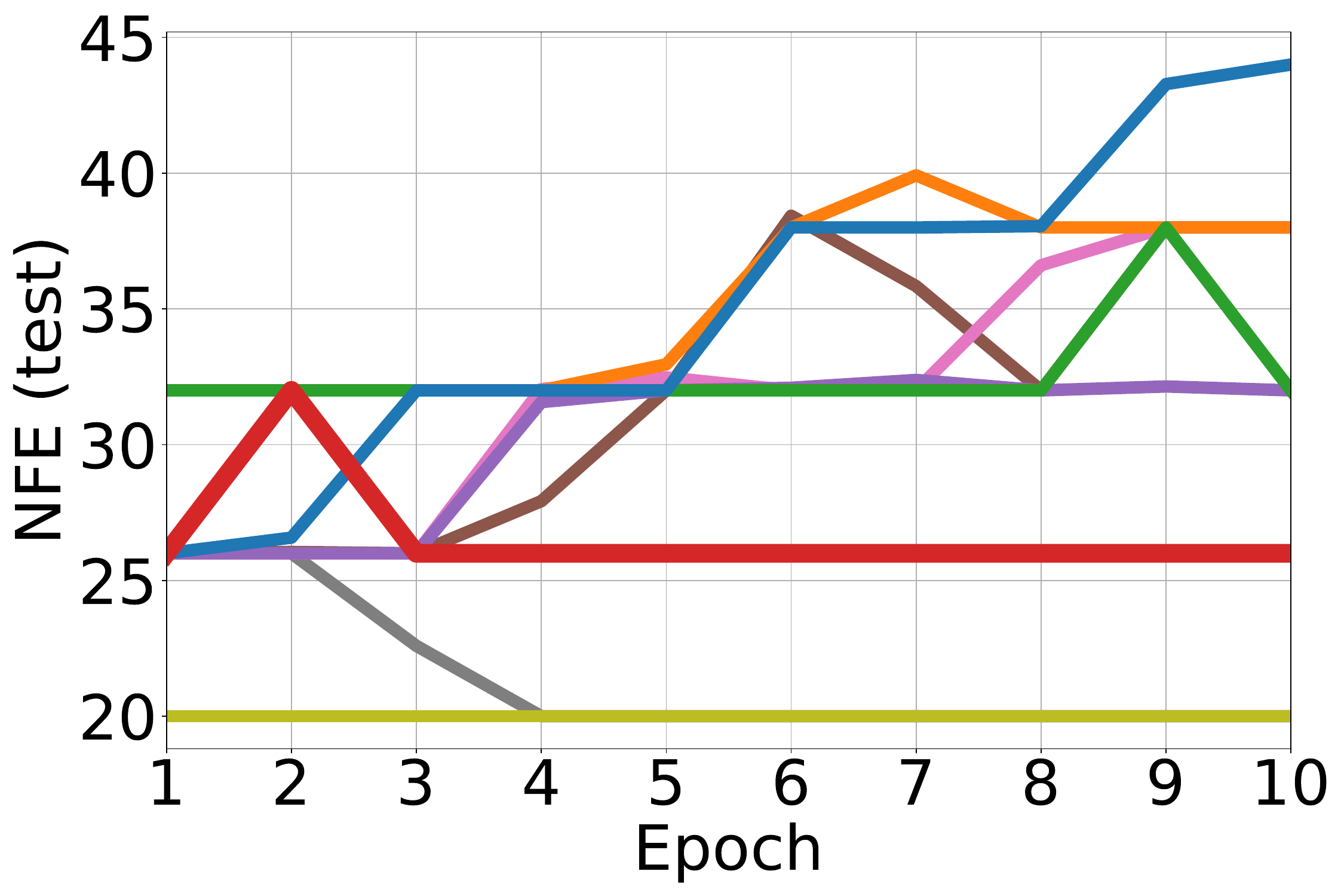}}
    
    \centering
    %
    % \vspace{-.5em}
    \caption{Train loss, train NFE (forward, backward), and test forward NFE on STL-10 Classification}
    % \vspace{-1.3em}
    \label{fig:IC_STL10}
\end{figure}

This section presents additional evaluation metrics (train loss, train forward NFE, train backward NFE, test forward NFE) for image classification on MNIST, CIFAR-10, CIFAR-100, and STL-10 datasets. In the image classification tasks on MNIST, CIFAR-10, and CIFAR-100 datasets, our BFNO-NODE shows the best performance in train forward NFE, train backward NFE, and test forward NFE compared to the baselines. In the image classification tasks on STL-10 dataset, our BFNO-NODE has the lowest value of train loss compared to the baselines, and in the case of NFE, we show superior results except for NODE and ANODE results. However, the test accuracy in Table~\ref{tab:Imageclassification_param_acc} is better than those of other baselines for the most of datasets.

\section{Feature Map Analyses}\label{a:tsne}
\begin{figure}[h]
    \centering
    \subfigure[][NODE]{\includegraphics[width=0.33\textwidth]{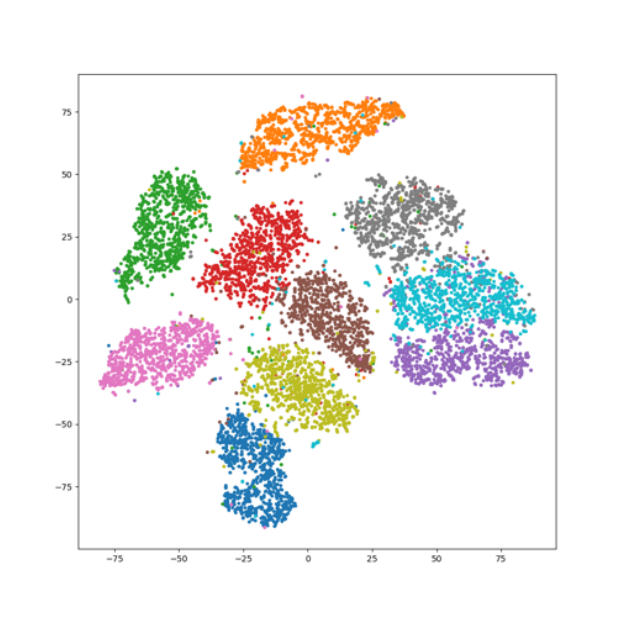}}
    \subfigure[][ANODE]{\includegraphics[width=0.33\textwidth]{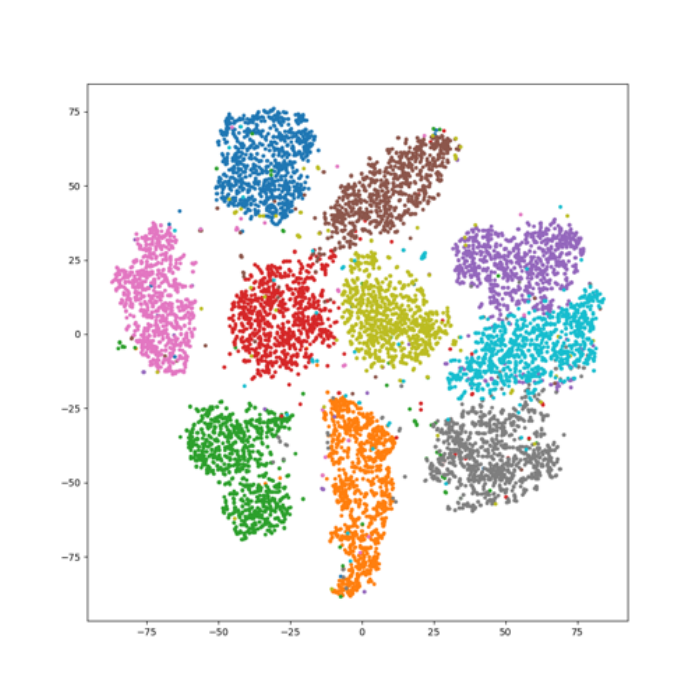}}
    \subfigure[][SONODE]{\includegraphics[width=0.33\textwidth]{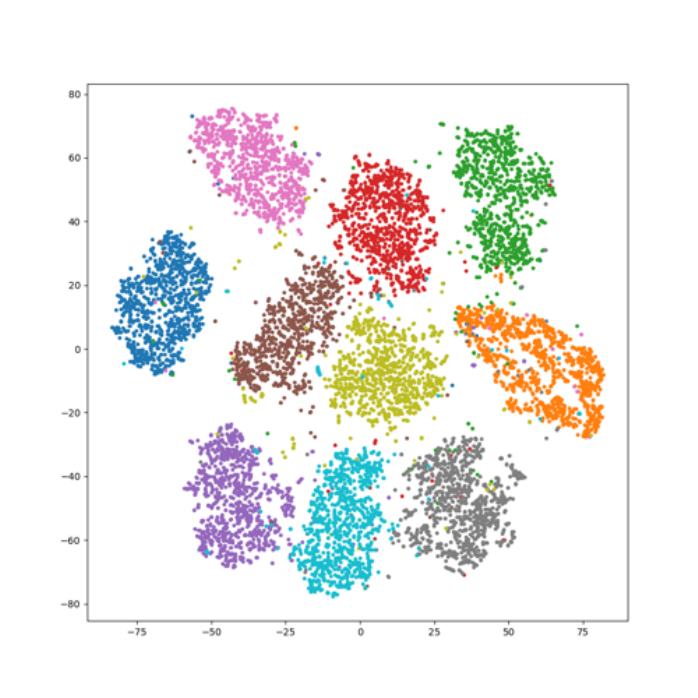}}
    \subfigure[][HBNODE]{\includegraphics[width=0.33\textwidth]{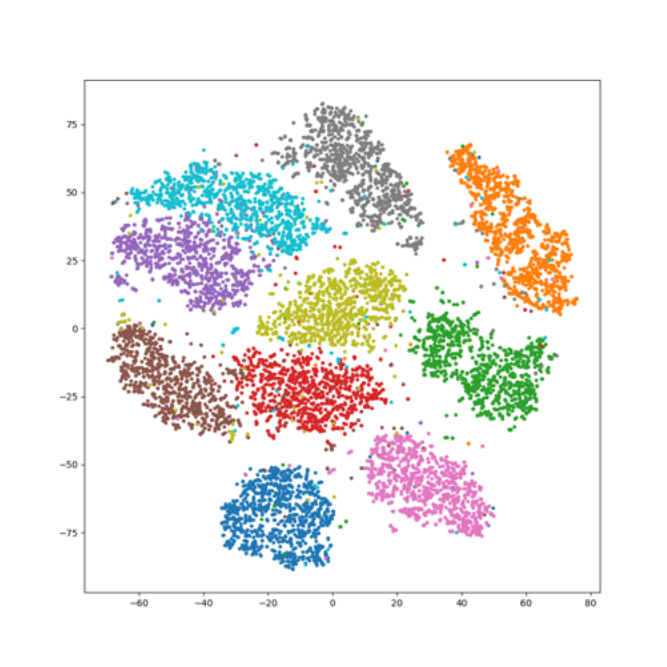}}
    \subfigure[][GHBNODE]{\includegraphics[width=0.33\textwidth]{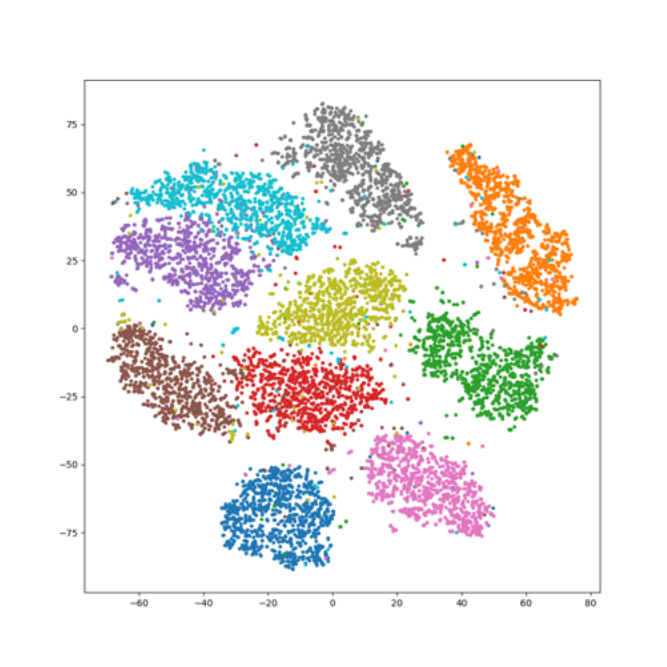}}
    \subfigure[][BFNO-NODE (ours)]{\includegraphics[width=0.33\textwidth]{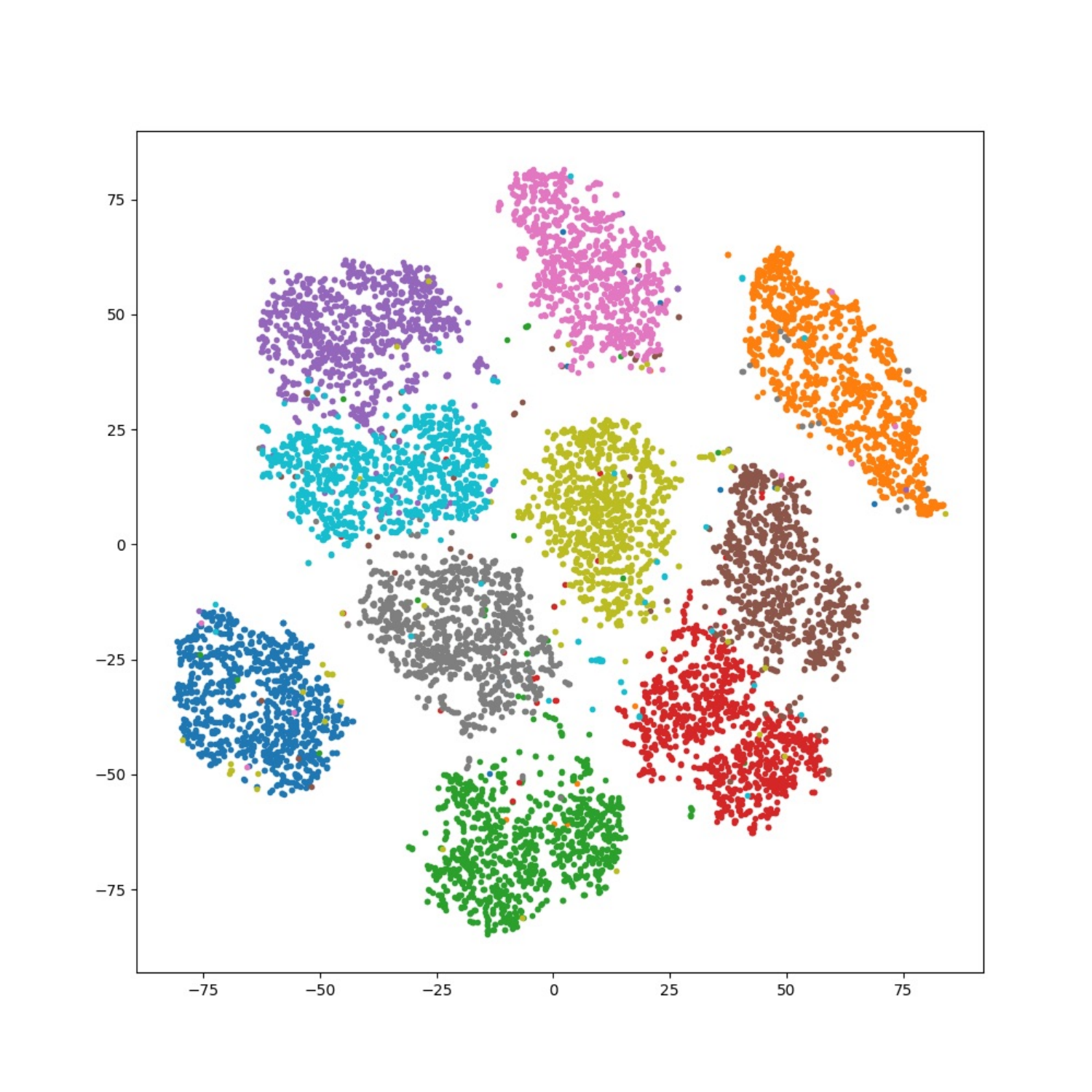}}
    % \vspace{-.5em}
    \caption{Feature map visualization for MNIST. The feature maps are projected into a two-dimensional space using t-SNE.}
    % \vspace{-.5em}
\end{figure}
These are the results of visualizing the feature maps of the NODE baselines and our BFNO-NODE when using MNIST. For the hidden vector before the last classification layer, it can be seen that our proposed BFNO-NODE classifies well in its hidden space.

\newpage

\section{Detailed Ablation Studies}
\label{sec:detail_ablation}
\begin{figure}[h]
    \centering
    \subfigure[][Ablation study on the neural operator type (MNIST)]{\includegraphics[width=0.32\textwidth]{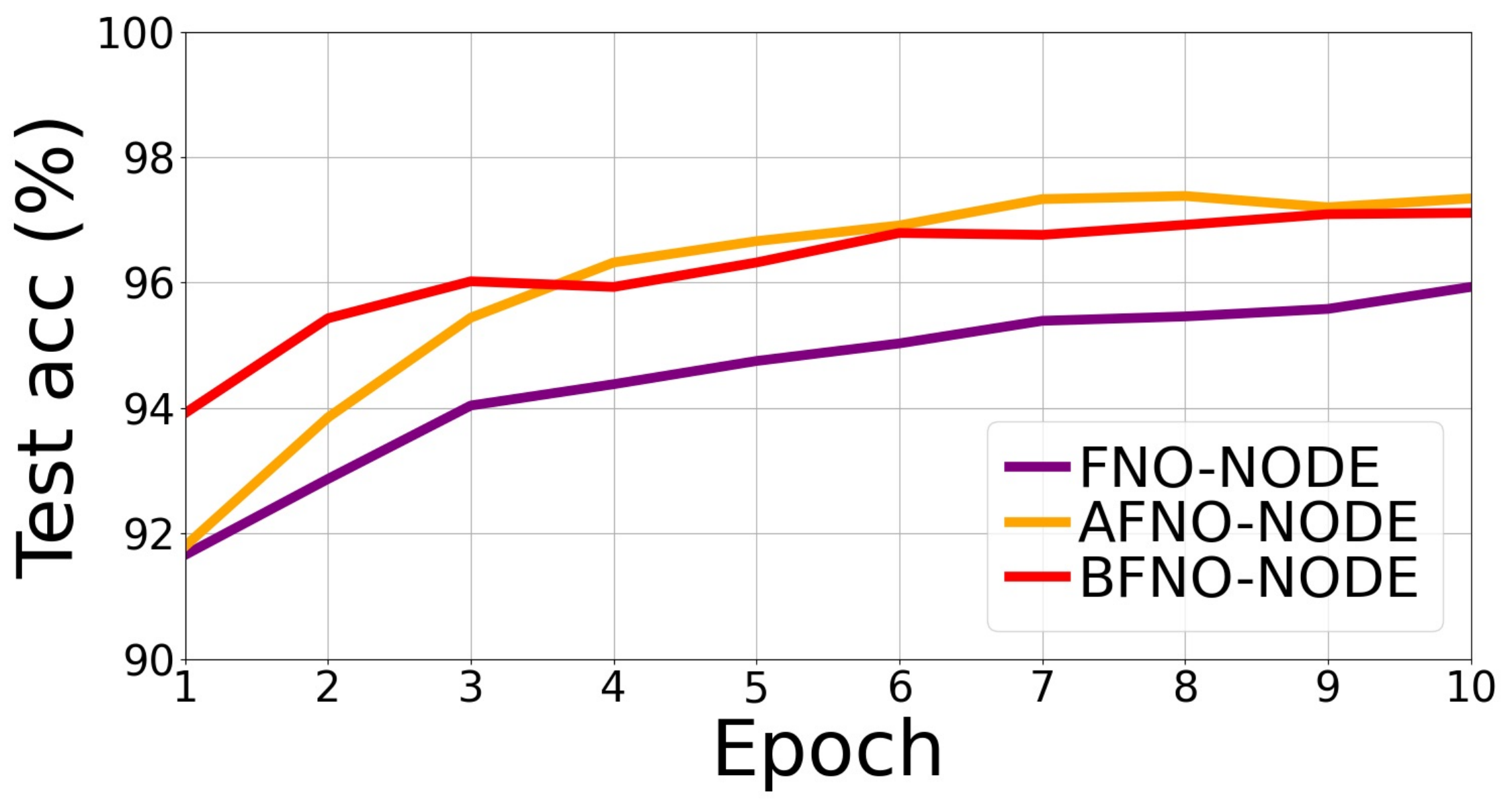}}
    \subfigure[][Ablation study on the neural operator type (CIFAR-10)]{\includegraphics[width=0.32\textwidth]{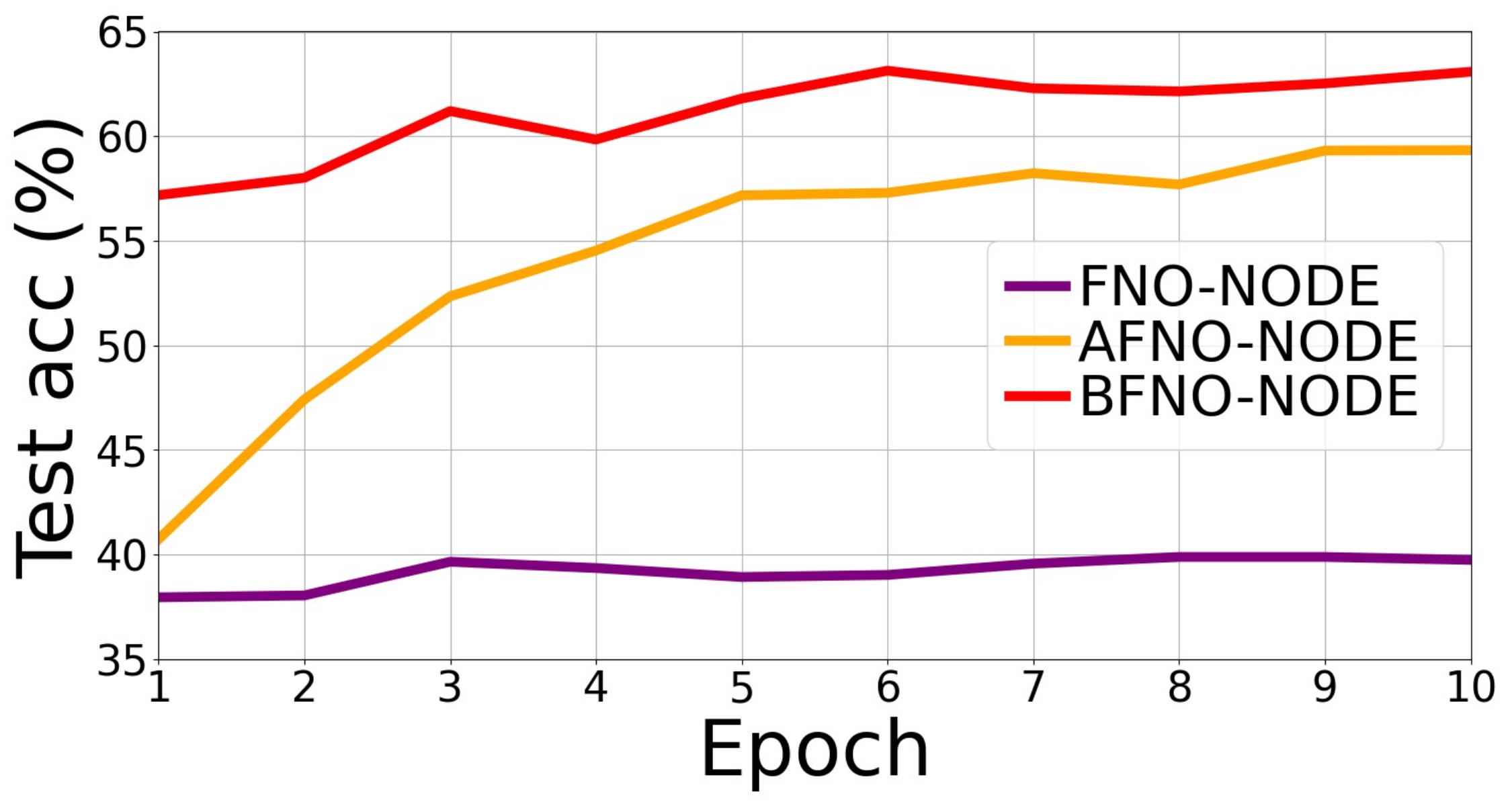}}
    \subfigure[][Ablation study on the neural operator type (CIFAR-100)]{\includegraphics[width=0.32\textwidth]{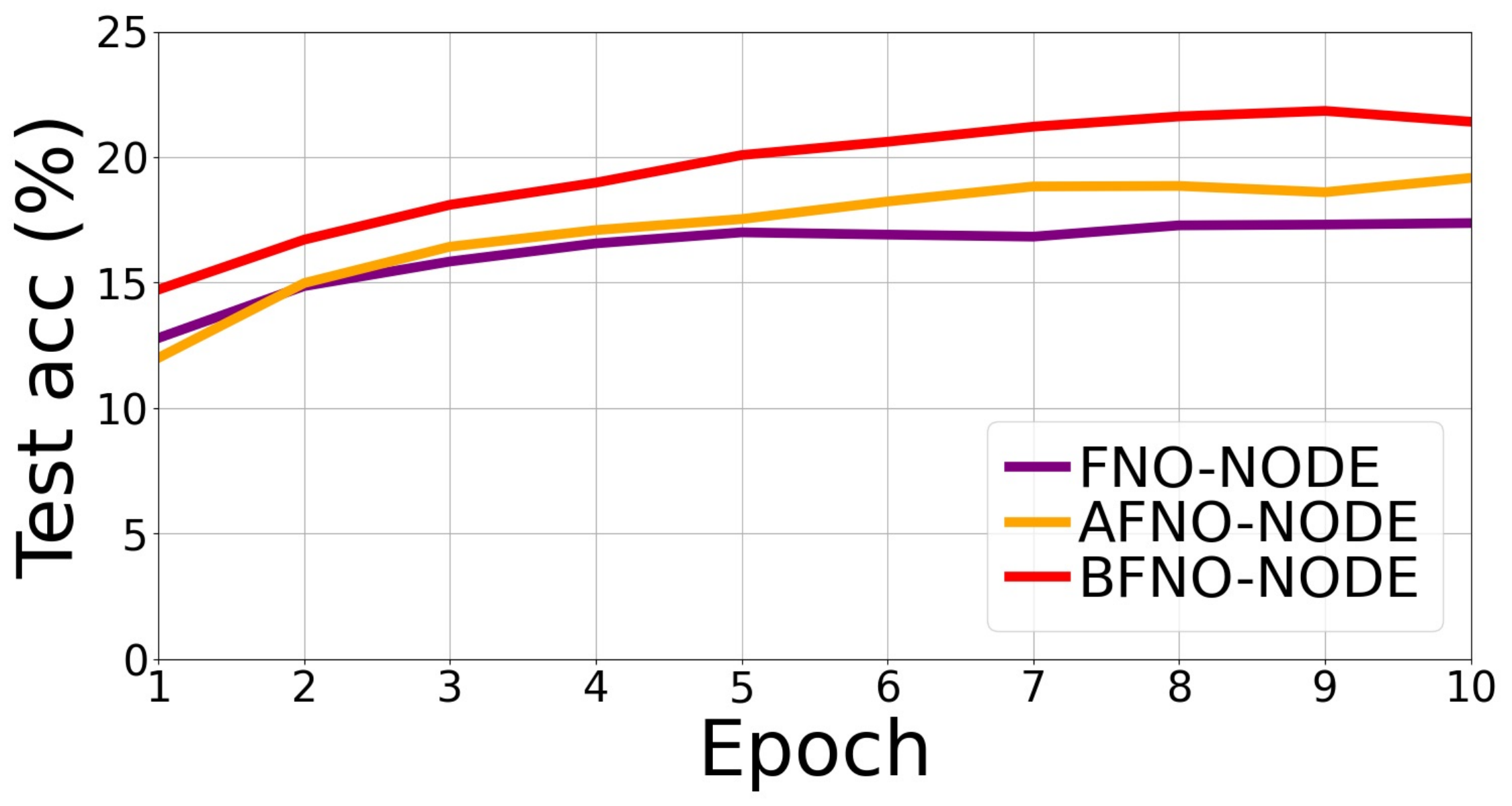}}
    % \vspace{-.5em}
    \caption{Ablation study on the two key design options of BFNO}
    % \vspace{-.5em}
\end{figure}

As shown in Figure~\ref{fig:ablation_study}, the performance of the BFNO structure as the ODE function $f$ is superior to that of the other two baselines, FNO and AFNO in STL-10. In this section, we compare the performance of the models for three additional datasets used in the image classification task. The experimental settings for FNO-NODE, AFNO-NODE, and BFNO-NODE are the same as those in Section~\ref{sec:image_classification}. However, AFNO-NODE has too long training time in CIFAR-100. Therefore, we contrast the performance of small-sized models (\# of parameters of FNO: 310,807, AFNO: 311,015, BFNO: 312,503).

\section{Detailed Settings for Image Classification Experiments}
\label{sec:detail_image_classification}

\subsection{Architecture of the ODE function $f$}
\begin{table}[h]
\centering
    \begin{minipage}{0.49\linewidth}
        \centering
        \small
        \renewcommand{\arraystretch}{1.2}
        % \vspace{-1em}
        \begin{tabular}{cccc}
        \specialrule{1pt}{1pt}{1pt}
        Layer & Design & Input size & Output size  \\ \hline
        1 & $Encoder$ & 2 $\times$ 28 $\times$ 28 & 47 $\times$ 28 $\times$ 28  \\
        2 & ReLU($BFNO$) & 47 $\times$ 28 $\times$ 28    & 47 $\times$ 28 $\times$ 28      \\ 
        3 & ReLU($BFNO$) &  47 $\times$ 28 $\times$ 28    &  47 $\times$ 28 $\times$ 28        \\ 
        4 & $BFNO$ &  47 $\times$ 28 $\times$ 28   &  47 $\times$ 28 $\times$ 28    \\  
        5 & $Decoder$ & 47 $\times$ 28 $\times$ 28  &  1 $\times$ 28 $\times$ 28   \\  
        \specialrule{1pt}{1pt}{1pt}
        \end{tabular}
        \caption{MNIST structure}
        \label{table:MNIST_structure}
    \end{minipage}
    \hfill
    \begin{minipage}{0.49\linewidth}
        \centering
        \small
        \renewcommand{\arraystretch}{1.2}
        % \vspace{-1em}
        \begin{tabular}{cccc}
        \specialrule{1pt}{1pt}{1pt}
        Layer & Design & Input size & Output size  \\ \hline
        1 & $Encoder$ & 4 $\times$ 32 $\times$ 32 & 76 $\times$ 32 $\times$ 32 \\
        2 & ReLU($BFNO$) & 76 $\times$ 32 $\times$ 32 & 76 $\times$ 32 $\times$ 32 \\ 
        3 & ReLU($BFNO$) & 76 $\times$ 32 $\times$ 32 & 76 $\times$ 32 $\times$ 32 \\ 
        4 & $BFNO$  & 76 $\times$ 32 $\times$ 32  & 76 $\times$ 32 $\times$ 32 \\  
        5 & $Decoder$ & 76 $\times$ 32 $\times$ 32 & 3 $\times$ 32 $\times$ 32 \\  
        \specialrule{1pt}{1pt}{1pt}
        \end{tabular}
        \caption{CIFAR-10 structure}
        \label{table:CIFAR10_structure}
    \end{minipage}
\end{table}

\begin{table}[ht!]
\centering
    \begin{minipage}{0.49\linewidth}
        \centering
        \small
        \setlength{\tabcolsep}{5pt}
        \renewcommand{\arraystretch}{1.2}
        % \vspace{-1em}
        \begin{tabular}{cccc}
        \specialrule{1pt}{1pt}{1pt}
        Layer & Design & Input size & Output size  \\ \hline
        1 & $Encoder$ & 4 $\times$ 32 $\times$ 32 & 118 $\times$ 32 $\times$ 32 \\
        2 & ReLU($BFNO$) & 118 $\times$ 32 $\times$ 32 & 118 $\times$ 32 $\times$ 32 \\ 
        3 & ReLU($BFNO$) & 118 $\times$ 32 $\times$ 32 & 118 $\times$ 32 $\times$ 32 \\ 
        4 & $BFNO$ & 118 $\times$ 32 $\times$ 32 & 118 $\times$ 32 $\times$ 32 \\  
        5 & $Decoder$ & 118 $\times$ 32 $\times$ 32 & 3 $\times$ 32 $\times$ 32 \\  
        \specialrule{1pt}{1pt}{1pt}
        \end{tabular}
        \caption{CIFAR-100 structure}
        \label{table:CIFAR100_structure}
    \end{minipage}
    \hfill
    \begin{minipage}{0.49\linewidth}
        \centering
        \small
        \renewcommand{\arraystretch}{1.2}
        % \vspace{-1em}
        \begin{tabular}{cccc}
        \specialrule{1pt}{1pt}{1pt}
        Layer & Design & Input size & Output size  \\ \hline
        1 & $Encoder$ & 4 $\times$ 96 $\times$ 96 & 99 $\times$ 96 $\times$ 96 \\
        2 & ReLU($BFNO$) & 99 $\times$ 96 $\times$ 96 & 99 $\times$ 96 $\times$ 96 \\ 
        3 & ReLU($BFNO$) & 99 $\times$ 96 $\times$ 96 & 99 $\times$ 96 $\times$ 96 \\ 
        4 & $BFNO$  & 99 $\times$ 96 $\times$ 96 & 99 $\times$ 96 $\times$ 96 \\  
        5 & $Decoder$ & 99 $\times$ 96 $\times$ 96 & 3 $\times$ 96 $\times$ 96 \\  
        \specialrule{1pt}{1pt}{1pt}
        \end{tabular}
        \caption{STL-10 structure}
        \label{table:STL10_structure}
    \end{minipage}
\end{table}

\subsection{Best Hyperparameters}
For each of the reported results, we list the best hyperparameter as follows:
\begin{itemize}
    \item For MNIST, learning rate = 0.001, relative tolerance = 0.001, absolute tolerance = 0.001, $L$ = 2, $N$ = 3; 
    \item For CIFAR-10, learning rate = 0.001, relative tolerance = 0.001, absolute tolerance = 0.001, $L$ = 2, $N$ = 3; 
    \item For CIFAR-100, learning rate = 0.001, relative tolerance = 0.001, absolute tolerance = 0.001, $L$ = 2, $N$ = 3; 
    \item For STL-10, learning rate = 0.001, relative tolerance = 0.001, absolute tolerance = 0.001, $L$ = 2, $N$ = 3; 
\end{itemize}

\section{Detailed Settings for Time Series Classification Experiments}
\label{sec:detail_time_series}

\subsection{Architecture of the ODE function $f$}
\begin{table}[h]
\centering
    \begin{minipage}{0.49\linewidth}
        \centering
        \small
        \renewcommand{\arraystretch}{1.2}
        \begin{tabular}{cccc}
        \specialrule{1pt}{1pt}{1pt}
        Layer & Design & Input size & Output size  \\ \hline
        1 & $Encoder$ & 15 $\times$ 3 $\times$ 50 & 500 $\times$ 3 $\times$ 50 \\
        2 & Tanh($BFNO$) & 500 $\times$ 3 $\times$ 50 & 500 $\times$ 3 $\times$ 50 \\ 
        3 & $Decoder$ & 500 $\times$ 3 $\times$ 50 & 15 $\times$ 3 $\times$ 50 \\   
        \specialrule{1pt}{1pt}{1pt}
        \end{tabular}
        \caption{Human Activity structure}
        \label{table:human_activity_structure}
    \end{minipage}
    \hfill
    \begin{minipage}{0.49\linewidth}
        \centering
        \small
        \renewcommand{\arraystretch}{1.2}
        \begin{tabular}{cccc}
        \specialrule{1pt}{1pt}{1pt}
        Layer & Design & Input size & Output size  \\ \hline
        1 & $Encoder$ & 15 $\times$ 3 $\times$ 50 & 50 $\times$ 3 $\times$ 50 \\
        2 & Tanh($BFNO$) & 50 $\times$ 3 $\times$ 50 & 50 $\times$ 3 $\times$ 50 \\ 
        3 & $Decoder$ & 50 $\times$ 3 $\times$ 50 & 15 $\times$ 3 $\times$ 50 \\  
        \specialrule{1pt}{1pt}{1pt}
        \end{tabular}
        \caption{PhysioNet structure}
        \label{table:physionet_structure}
    \end{minipage}
\end{table}
\subsection{Best Hyperparameters}
For each of the reported results, we list the best hyperparameter as follows:
\begin{itemize}
    \item For Human Activity, learning rate = 0.0001, relative tolerance = 0.001, absolute tolerance = 0.0001, $L$ = 1, $N$ = 1. 
    \item For PhysioNet, learning rate = 0.0001, relative tolerance = 0.001, absolute tolerance = 0.0001, $L$ = 1, $N$ = 1. 
\end{itemize}

\section{Detailed Settings for Image Generation Experiments}
\label{sec:detail_image_generation}

\subsection{Architecture of the ODE function $f$}
\begin{table}[ht!]
\centering
    \begin{minipage}{0.49\linewidth}
        \centering
        \small
        \setlength{\tabcolsep}{5pt}
        \renewcommand{\arraystretch}{1.2}

        \begin{tabular}{cccc}
        \specialrule{1pt}{1pt}{1pt}
        Layer & Design & Input size & Output size  \\ \hline
        1 & $Encoder$ & 2 $\times$ 28 $\times$ 28 & 100 $\times$ 28 $\times$ 28 \\
        2 & SoftPlus($BFNO$) & 100 $\times$ 28 $\times$ 28 & 100 $\times$ 28 $\times$ 28 \\ 
        3 & SoftPlus($BFNO$) & 100 $\times$ 28 $\times$ 28 & 100 $\times$ 28 $\times$ 28 \\ 
        4 & SoftPlus($BFNO$) & 100 $\times$ 28 $\times$ 28 & 100 $\times$ 28 $\times$ 28 \\
        5 & $Decoder$ & 100 $\times$ 28 $\times$ 28 & 1 $\times$ 28 $\times$ 28 \\  
        \specialrule{1pt}{1pt}{1pt}
        \end{tabular}
        \label{table:generation_mnist_structure}
        \caption{MNIST structure}
    \end{minipage}
    \hfill
    % \vspace{.5em}
    \begin{minipage}{0.49\linewidth}
        \centering
        \small
        \setlength{\tabcolsep}{5pt}
        \renewcommand{\arraystretch}{1.2}
        \begin{tabular}{cccc}
        \specialrule{1pt}{1pt}{1pt}
        Layer & Design & Input size & Output size  \\ \hline
        1 & $Encoder$ & 4 $\times$ 32 $\times$ 32 & 100 $\times$ 32 $\times$ 32 \\
        2 & SoftPlus($BFNO$) & 100 $\times$ 32 $\times$ 32 & 100 $\times$ 32 $\times$ 32 \\ 
        3 & SoftPlus($BFNO$) & 100 $\times$ 32 $\times$ 32 & 100 $\times$ 32 $\times$ 32 \\ 
        4 & SoftPlus($BFNO$) & 100 $\times$ 32 $\times$ 32 & 100 $\times$ 32 $\times$ 32 \\  
        5 & $Decoder$ & 100 $\times$ 32 $\times$ 32 & 3 $\times$ 32 $\times$ 32 \\  
        \specialrule{1pt}{1pt}{1pt}
        \end{tabular}
        \caption{CIFAR-10 structure}
        \label{table:generation_CIFAR10_structure}
    \end{minipage}
\end{table}
\subsection{Best Hyperparameters}
For each of the reported results, we list the best hyperparameter as follows:
\begin{itemize}
    \item For MNIST, learning rate = 0.0001, relative tolerance = 0.001, absolute tolerance = 0.001, $L$ = 3, $N$ = 1. 
    \item For CIFAR-10, learning rate = 0.0001, relative tolerance = 0.001, absolute tolerance = 0.001, $L$ = 3, $N$ = 1. 
\end{itemize}

\end{document}